\documentclass[10pt,twocolumn,letterpaper]{article}

\usepackage[pagenumbers]{cvpr} 

\usepackage{times}
\usepackage{epsfig}
\usepackage{graphicx}
\usepackage{amsmath}
\usepackage{amssymb}
\usepackage{booktabs}

\usepackage[pagebackref,breaklinks,colorlinks]{hyperref}

\newcommand{\COMMENT}[1]{}

\usepackage[capitalize]{cleveref}
\crefname{section}{Sec.}{Secs.}
\Crefname{section}{Section}{Sections}
\Crefname{table}{Table}{Tables}
\crefname{table}{Tab.}{Tabs.}

\def\cvprPaperID{8023} 
\def\confName{CVPR}
\def\confYear{2023}

\begin{document}

\title{A Light-Weight Contrastive Approach for \\ Aligning  Human Pose Sequences}

\author{Robert T.~Collins\thanks{This work originated while on sabbatical from Penn State and visiting Carnegie Mellon University, hosted by Professor Martial Hebert.}\\
Penn State University\\
{\tt\small rtc12@psu.edu}
}
\maketitle

\begin{abstract}
We present a simple unsupervised method for learning an encoder mapping short 3D pose sequences into embedding vectors  suitable for sequence-to-sequence alignment by dynamic time warping.  Training samples consist of temporal windows of frames  containing 3D body points such as mocap markers or skeleton joints.  A light-weight, 3-layer encoder is trained using a contrastive loss function that encourages embedding vectors of augmented sample pairs to have cosine similarity 1, and similarity 0 with all other samples in a minibatch.  When multiple scripted training sequences are available, temporal alignments inferred from an initial round of training are  harvested to extract additional, cross-performance match pairs for a second phase of training to refine the encoder.  In addition to being simple, the proposed method is fast to train, making it easy to adapt to new data using different marker sets or skeletal joint layouts.  Experimental results illustrate ease of use, transferability, and utility of  the learned embeddings for comparing and analyzing human behavior sequences.
\end{abstract}

\section{Introduction}
\label{sec:intro}

In  pursuit of human behavior understanding, there is frequent need for aligning sequences of 3D point data representing  body pose over time. Typically, this point data represents either directly measured motion capture (mocap) marker locations on the exterior of the body, or estimated joint locations forming an interior skeletal structure.  We address both types of human 3D point data in this work.

\begin{figure}[b]
  \centering  \hfill
  \begin{subfigure}{0.98\linewidth}
    \includegraphics[width=1.0\linewidth]{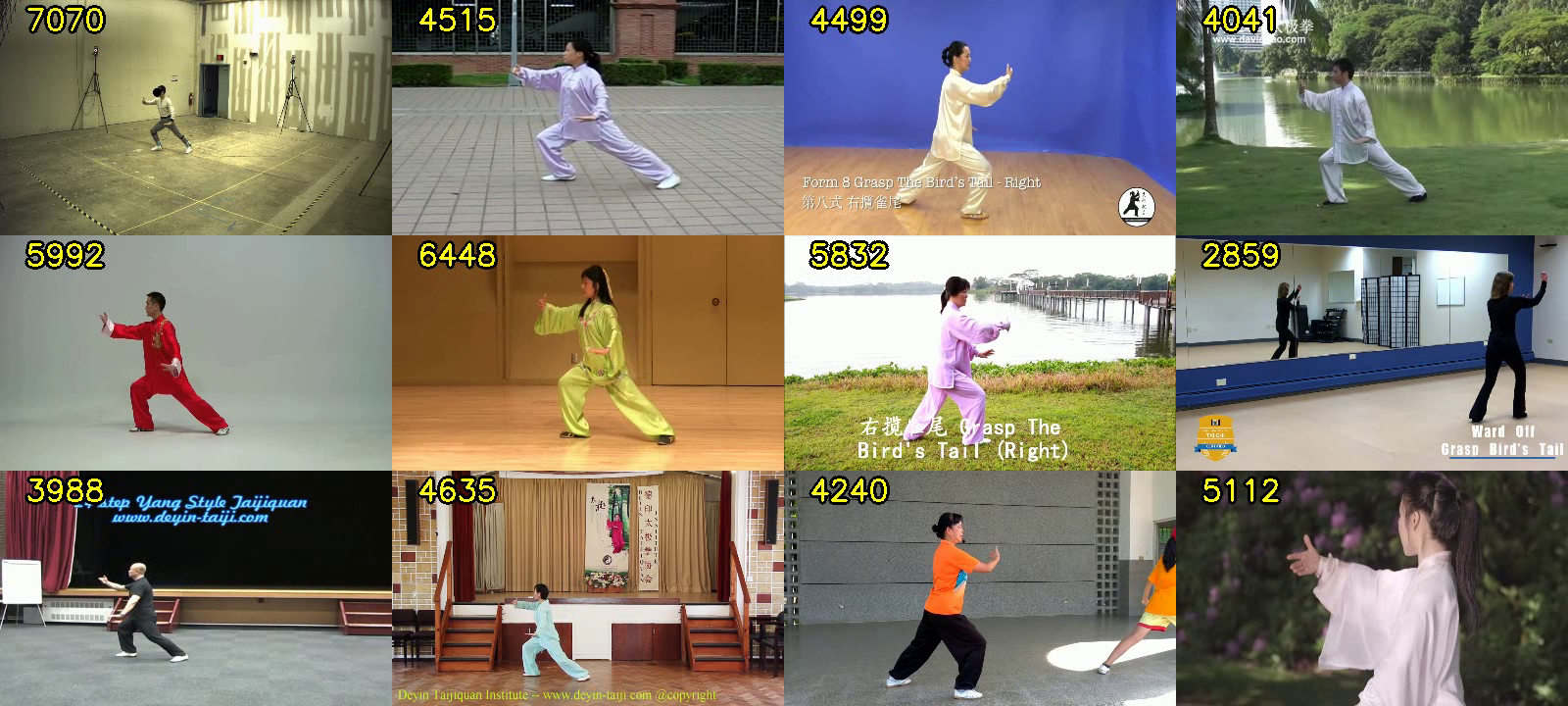}
  \end{subfigure} \hfill
  \caption{One outcome of this work is aligning ``in the wild"  Youtube videos to a reference  performance captured in a mocap lab (upper left). See additional frames in the Appendix.}
  \label{fig:youtubepics}
\end{figure}

Temporal alignment of pose sequences can contribute to several tasks in  human motion understanding. 
1) Determining correspondence between the same or substantially similar poses facilitates empirical study of the range of variability in performing them. 
2) Detecting repeated actions, either short term periodic motions that occur with a frequency, or actions that repeat after longer gaps in time, identifies potential ``words" in a vocabulary of action units.
3) Transfering labels from a reference performance to other performances that have been aligned with it, similar to the way a labeled ``atlas" is used in medical imaging, can locate special keyposes or transfer a temporal segmentation and labeling of actions onto a new sequence.
4) Detecting misalignments identifies performance anomalies such as missing actions or out of order performance of actions in a sequence.

In this paper we present a simple and light-weight contrastive method for unsupervised learning of an encoder that maps short pose sequences into an embedding space.  Two pose sequences can then be compared using cosine similarity of their embedding vectors.  When used in conjunction with samples extracted in temporal order, 
the pairwise cosine distance matrix is a suitable cost matrix for doing sequence-to-sequence alignment by dynamic time warping.  This alignment can then be mined to extract cross-performance matching pairs, and these  can be included during a second phase of training to refine the encoder.
We show that the method is effective at producing discriminative representations leading to high quality sequence-to-sequence alignments (Figure~\ref{fig:youtubepics}).
In addition to being simple, the method is fast to train, making it easy to adapt to new data that uses different marker sets or skeletal joint layouts.

\section{Related Work}

\subsection{Pose Sequence Representations}

Most approaches in the literature for representing sequences of body poses as a sparse set
of 3D points observed over time fall into graph-based versus image-based representations.  Graph-based models explicitly represent the connectivity of points in a skeleton, and are typically used with graph convolutional network (GCN) architectures \cite{yan2018STGCN,rethinkingSTGCN,Shi2019TwoStreamAG}.  This approach introduces a strong architectural prior of spatial locality that, we feel, imposes an unnecessarily strict structural, inductive bias.  Body segments not directly connected in a tree structured graph may nonetheless be highly correlated in their movements, such as an arm swinging in synchrony with the opposite leg.

An alternative is image-based representations that  concatenate point coordinates into one or more 2D image arrays  \cite{Du2015,Ke2017,SkeleMotion}, including heat map images \cite{revisitingSkeletons}, that can be processed by a normal feedforward CNN or RNN architecture.  We adopt this approach; our input data structure is a  2D array of 3D point coordinates (across rows) seen at different time steps (down columns).  Although our pose input arrays can be thought of as an image, they are less image-like than some works \cite{SkeleMotion} and 
 we prefer to think of our 2D array as a set of point coordinate trajectories.  Since there is no implicit connectivity bias, our CNN encoder  can learn correlations between arbitrary body points from the data, and we can swap in datasets with different skeletal topologies or permutations of points with little or no modification. 

\vspace{-1ex}
\subsection{Learning Encoders by Siamese Networks}

Siamese networks have long been used for unsupervised representation learning.
Early efforts made use of explicit positive (similar) pairs and negative (dissimilar) pairs,
and learned an encoder using a contrastive loss function encouraging small  distance
between vectors representing positive pairs while maintaining a larger distance (margin) 
between negative pairs \cite{HadsellEarlySiamese}.
Recent work \cite{SimCLR,BYOL,SimSiam,MoCo,SWaV} seeks to learn representations with positive pairs alone, often by transforming a single training example into a positive pair through augmentation.
This approach allows generation of  a potentially unlimited number of positive pairs;
however, naive use of only positive pairs leads to collapsing solutions such as  all examples
 mapping to a constant vector.  
 The current research trend is to avoid collapse by tweaking either the loss or the architecture
 to reintroduce (implicitly) contrasting negatives \cite{SimCLR, MoCo}, to enforce 
formation of multiple distinct clusters in  embedding space \cite{SWaV}, or to build 
operational asymmetries into the Siamese architecture \cite{SimSiam,BYOL}.

The Barlow Twins (BT) architecture \cite{BarlowTwins} introduces a novel least-squares loss function comparing an empirical correlation matrix to an ideal identity matrix.  Their loss  promotes decorrelation of learned embedding feature components and prevents collapse without asymmetric learning elements; a BT network uses no predictor, stop-grad, momentum, or clustering.  We use a similar least-squares loss function but with an important difference: we compare a matrix of cosine similarities between embedding vectors to the identity matrix.  This seemingly small difference leads to a very different interpretation (\Cref{sec:lossfunction}).

\vspace{-1ex}
\subsection{Sequence-to-Sequence Temporal Alignment}

Nonlinear but monotonic alignment of two time series is elegantly addressed by Dynamic Time Warping (DTW), an efficient
algorithm based on dynamic programming \cite{TimeAlignPatRec}.  However, success or failure
crucially depends on how discriminative the pairwise similarity function is.
 DTW using L2 distance can fail even on scalar-valued signals when amplitudes differ, 
 causing singularities where a single point in one curve maps to large sections of another, motivating methods that incorporate more local context such as derivatives \cite{DDTW}.
For multivariate time series data, methods like Canonical Time Warping (CTW) \cite{CTW} combine DTW with multivariate analysis techniques that do feature selection and weighting among the data components. 
Scott et.al.~\cite{scott20174d} use CTW to align 2D joints extracted from video with projected 3D mocap joints to iteratively estimate both the spatial alignment (camera viewpoint) and temporal alignment of scripted Taiji sequences.  
More recently, Deep CTW replaces linear projection matrices with nonlinear features learned by a deep network \cite{DeepCTW}.  

Other recent deep network alignment approaches operate directly on raw video, typically using an initial ResNet-50 backbone for feature extraction \cite{Dwibedi_2019_CVPR, Hadji_2021_CVPR, Haresh_2021_CVPR, Liu_2022_CVPR}.   These  approaches are only demonstrated on short videos (a few seconds) containing single actions, e.g.~from the Penn Action Dataset \cite{Penn_Action_Dataset}.  In contrast, we present experiments on five minute long, complex pose sequences containing multiple actions. 
CASA \cite{Kwon_CVPR_2022}, which is similarly focused on unsupervised learning to align 3D skeleton sequences,  uses a more complicated transformer architecture with self- and cross-attention layers and a more complicated domain-specific set of data augmentations.  
Comparisons with CASA and other works on the Penn Action Dataset are presented in Section~\ref{sec:PennAction}.

\section{Learning an Encoder for Pose Sequences}

In this section we present a simple, contrastive method for unsupervised learning of an encoder that maps short pose sequences into embedding vectors.  An input  sequence is a temporal window of poses sampled from a longer performance.  It is formatted as a 2D array of 3D points (across rows) at different time steps (down columns).  A row can be thought of as an ordered sequence of 3D points, or, alternatively, each column is a trajectory of  X, Y, or Z coordinates of one 3D point.   
The point order is arbitrary, e.g.~there is no need for a shoulder point to be adjacent to an elbow point, but it is important that whatever permutation of ordering there is remains consistent across training and testing.

Input data points are normalized using a domain-specific procedure specific to human body point data, and data augmentation transforms each input sample  into a positive pair of samples that the encoder learns to map to similar vectors.  Data normalization and augmentation are described when discussing training, in \Cref{sec:training}.   We first focus on the encoder architecture and contrastive loss function.

\subsection{Encoder Architecture}

\begin{figure}[t]
  \centering
   \includegraphics[width=0.99\linewidth]{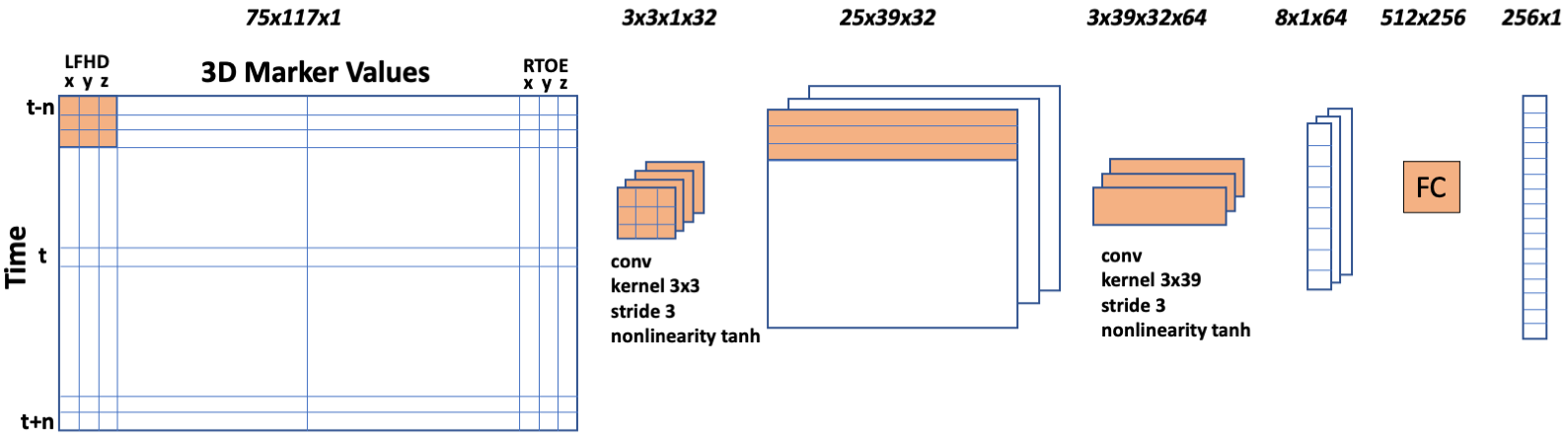}
   \caption{Shallow network for encoding temporal windows of 3D poses into embedding vectors,
   instantiated here for 39 mocap markers (3x39=117) viewed over a 3 second, 25 frame per second temporal window (3x25=75).}
   \label{fig:shallownetwork}
\end{figure}

We design a shallow, 3-layer network architecture to encode temporal samples into embedding vectors.  A sample design is depicted in \Cref{fig:shallownetwork}.  Each temporal window of data is input to the encoding network as an N x 3P array, where N is the number of frames, e.g.~75 for a 3 second temporal window sampled at 25 fps, and P is the number of 3D body points, e.g.~39 for a typical Vicon Plug-in Gait mocap marker set \cite{PlugInGaitReference}.  Adjacent sets of 3 columns contain X, Y and Z values for each marker.  The first network layer applies a bank of convolution filters focusing on describing short-term motion of individual markers.  Each filter has 3 columns corresponding to marker X, Y, Z values, and is applied with stride 3 so that it only considers X, Y, Z of each marker in isolation, specifically NOT mixing X,Y, Z values across different markers.  The intent is that filters in this layer will encode location and direction of motion of individual body points.  The second layer of filters spans across all markers spatially,  thus combining marker information across the entire pose, although again over only a short time window. This layer, hypothetically, can discover correlations in relative location and motion of  different body parts, including ones not directly connected in a skeletal tree structure.  A final fully connected layer combines all feature channels across the entire temporal window to produce a single output encoding vector.  This final 256x1 embedding vector encodes both pose and motion of the body across the input temporal sequence window.

\subsection{Loss Function\label{sec:lossfunction}}

\begin{figure}
  \centering
   \includegraphics[width=0.9\linewidth]{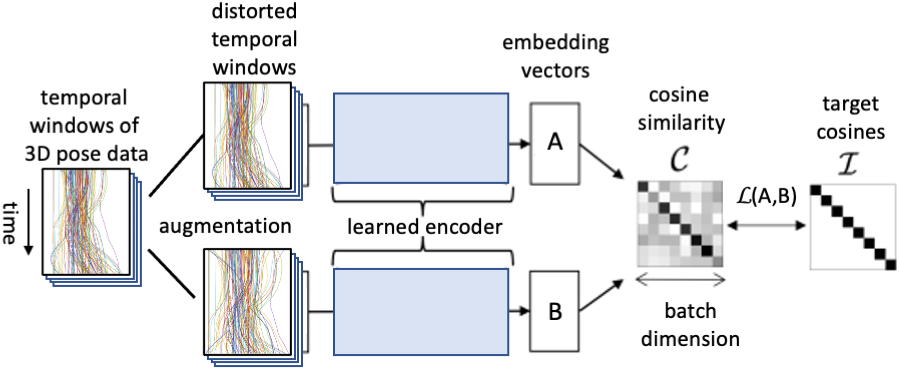}
   \caption{Siamese network for learning parameters of the encoder (top and bottom encoders share parameters).  
   }
   \label{fig:siamesenetwork}
\end{figure}

Parameters of the encoder are learned by a Siamese network architecture  (\Cref{fig:siamesenetwork}).
Similarity of embedding vectors will be measured by cosine similarity, the dot product of two vectors after scaling them to be unit vectors.  Given embedding vectors $a_i$ and $b_j$, their cosine similarity ${\cal C}$ is defined as
\begin{equation}
{\cal C}(a_i,b_j) = \frac{a_i^T b_j}{\| a_i\|_2 \; \| b_j\|_2} 
\end{equation}
where $\| \cdot \|_2$ denotes $l_2$-norm.  This score ranges from -1 to 1, with 1 meaning most similar.
A numeric evaluation of alternative similarity functions based on Euclidean distance and Hadsell (margin) distance~\cite{HadsellEarlySiamese} is included in the Appendix.

We would like to train the encoder network to map temporal windows containing similar pose and motion into embedding vectors that have pairwise cosine similarity of 1.  This will promote them forming a tight cluster on the unit hypersphere.  In contrast, we want two temporal windows containing different poses/motions to be mapped into embedding vectors with pairwise similarity of 0, forcing their representation vectors to be orthogonal.  We choose 0 rather than -1 as the target  score for mismatching vectors because we want  to prevent multiple vectors that don't match $a$, but also don't match each other, from clustering at the antipodal point of the hypersphere opposite from $a$.

Given a minibatch of $N$ positive training pairs of $D \times 1$ column vectors $\{(a_1,b_1),(a_2,b_2),...,(a_N,b_N)\}$, where $D$ is the dimension of the embedding space, e.g.~256, we collect the $a$ and $b$ vectors respectively as columns of $D \times N$ matrices $A$ and $B$ and
write the loss function as
\begin{equation}
{\cal L}(A,B) =   \sum_i ({\cal C}(a_i,b_i) - 1)^2 + \lambda  \sum_i  \sum_{j\neq i} {\cal C}(a_i,b_j)^2
\label{eq:ourlossfun}
\end{equation}
This is essentially a weighted sum of squares loss function computed between pairwise cosine similarities of encoding vectors and the desired values 1 or 0.
 In practice, we set $\lambda$ to a fixed value $1/(N-1)$ to evenly weight the $N$ diagonal terms with the  $N\mathord{*}(N-1)$ off-diagonal terms.

This loss equation has the same form as the Barlow Twins (BT) loss function (equation 1 in \cite{BarlowTwins}).  However, the meaning of ${\cal C}()$ is different between the two, leading to a very different interpretation.  To explore this in more detail, 
let  $(\hat{A}_{\downarrow})$ represent the operation of $l_2$-normalizing each column of A, and  $(\hat{A}_{\rightarrow})$ be $l_2$-normalization of each row.  If $\lambda=1$, our loss function could be written compactly as
\begin{equation}
{\cal L}(A,B) =   \| (\hat{A}_{\downarrow})^T (\hat{B}_{\downarrow}) - I_{NxN} \|_F^2   
\end{equation}
which is the squared Frobenius norm of the difference between $A^T B$ (after normalizing down the columns of each) and the $N \times N$ identity matrix.
Minimizing Frobenius norm is equivalent to minimizing sum of squares residual error.
In comparison, the loss function for a BT network is
\vspace{-1ex}
\begin{equation}
{\cal L_{BT}}(A,B) =   \| (\hat{A}_{\rightarrow}) (\hat{B}_{\rightarrow})^T - I_{DxD} \|_F^2   
\end{equation}
which is normalizing along the batch dimension instead.

This seemingly small difference gives the methods a totally different interpretation.
Pairwise terms  $A B^T$ in BT are interpreted as correlations, and the objective
function seeks to decorrelate the $D$ component dimensions of the feature embedding.
Our pairwise terms $A^T B$ are interpreted as cosine similarities, and our loss encourages embedding vectors of $N$ positive training pairs to have cosine similarity 1 with each other and 0 with all other pairs.  Our interpretation of loss is thus more like that of the common InfoNCE measure \cite{SimCLR} while being easier to compute as it is essentially a least squares loss.
Like infoNCE, and unlike BT, our approach is still a contrastive loss scheme; we don't have explicitly labeled negative examples, but within each minibatch, embedding vectors $a_i, b_j$ for $j \neq i$ are implicitly assumed to be negative pairs.  On the other hand, our method shares the same simplicity as BT; we don't need asymmetric  elements like prediction networks, momentum encoders, non-differentiable clustering operators, or stop-gradients to avoid collapsing solutions \cite{BarlowTwins}.

\subsection{Two Phases of Training \label{sec:training}}

The encoder is trained in two phases, each phase distinguished by
its source of  positive example training pairs.  In Phase 1, positive pairs are produced by
data augmentation, to be described below.  In Phase 2, an initial encoder learned in the first phase
is used to compute embedding vectors of uniformly
sampled temporal windows from different performances. 
For scripted sequences (same activities in same order), 
pairwise cosine distances between embedding vectors from 
 two different performances form a cost matrix suitable for dynamic time warping alignment. 
 The non-linear DTW path identifies matching
 temporal windows between two different performances, and these matches  are added as positive pairs to the training set and used to refine the encoder.
Harvested cross-performance pairs exhibit far greater and more realistic variability than the initial
data augmentation generator, allowing the refined encoder to better discriminate between positive and negative pairs of pose sequences, leading to more well-defined DTW cost matrices. 

\vspace{-1ex}
\subsubsection{Data Augmentation}

During training, each data sample is paired with an augmented sample (alternatively, two augmented samples could act as a training pair).  The augmented sample is generated by  extracting a new,  overlapping sample window from the pose sequence using a random temporal offset and scale.  Given a temporal window centered at time $t_0$ and sampled at a rate of $r$ frames per second, a new window is randomly sampled at temporal location  $t_0 + \delta t$ with sampling rate $(1+ \delta r)*r$.  In the current implementation with 3 sec windows, temporal offset $\delta t$ is uniformly distributed between -.5 and .5 sec and relative time scale $\delta r$ is uniformly distributed between -1/3 and 1/3.  This augmentation  jitters the temporal location of a sample window while simulating performing the movements faster or slower.

\vspace{-2ex}
\subsubsection{Data Normalization\label{sec:dataNormalize}}

Before training,  temporal windows of body pose data are normalized by centering, rotating and scale normalizing the data values.  Centering and rotation use a body-specific procedure to make the 3D point data invariant to the body's location and facing direction in the capture space. Processing to remove location and orientation variability has  been used in prior work \cite{predictandcluster}, and to save space we present our version of normalization in the Appendix.

\subsection{Implementation}

All code is implemented in Matlab, using Matconvnet 1.0-beta24 \cite{Matconvnet}.
The Siamese network  is implemented as a directed acycle graph (dag),
and training is performed using  cnn\_train\_dag, a stock backpropagation training
driver using stochastic gradient descent with momentum.  Phase 1 training is done for 15 epochs
using a batch size of 128 and a learning rate of 0.01. Phase 2 training incorporates cross-performance matches harvested by DTW sequence
matching using the initial encoder from Phase 1, and is performed for 30 epochs.
All training was done on a modest platform: a Mid-2012 Macbook Pro laptop
using CPU-only mode.  

\section{Experiments}

\subsection{PSU-TMM100 Dataset\label{sec:tmm100dataset}}

The Penn State Taiji MultiModal (PSU-TMM100) dataset  \cite{PSU-TMM100} contains 100 sequences of 24-form Yang Style Taiji, performed by 10 subjects with varying experience levels.  Each performance includes simultaneously recorded Vicon 3D marker and joint data, two synchronized and calibrated video views, and foot pressure maps measured by Tekscan F-scan insole sensors. In our experiments we only use the Vicon 3D marker and joint data.  The dataset can be requested from \url{http://vision.cse.psu.edu/data/data.shtml}.  It has a custom-written license.  The dataset was used previously in \cite{PSU-TMM100} to train algorithms to predict foot pressure from body pose.  

To train an encoder for aligning performances based on mocap marker data, we sampled temporal windows
from the first two ``takes" (performances) of subjects 1 through 9 (18 takes), leaving out subject 10 to test generalization to new subjects.  Overlapping temporal windows were extracted uniformly at 0.5 second intervals with each window being 3 seconds long and subsampled to 25 fps.  Data was normalized and augmented to form positive training pairs as described in the last section.  After an initial Phase 1 encoder was trained, it was used with DTW to time align all 18-choose-2 
cross-performance takes.  Additional positive training pairs were harvested from these alignments and combined with the initial training pairs to refine the encoder in Phase 2.

\begin{figure}[t]
  \centering
   \includegraphics[width=1\linewidth]{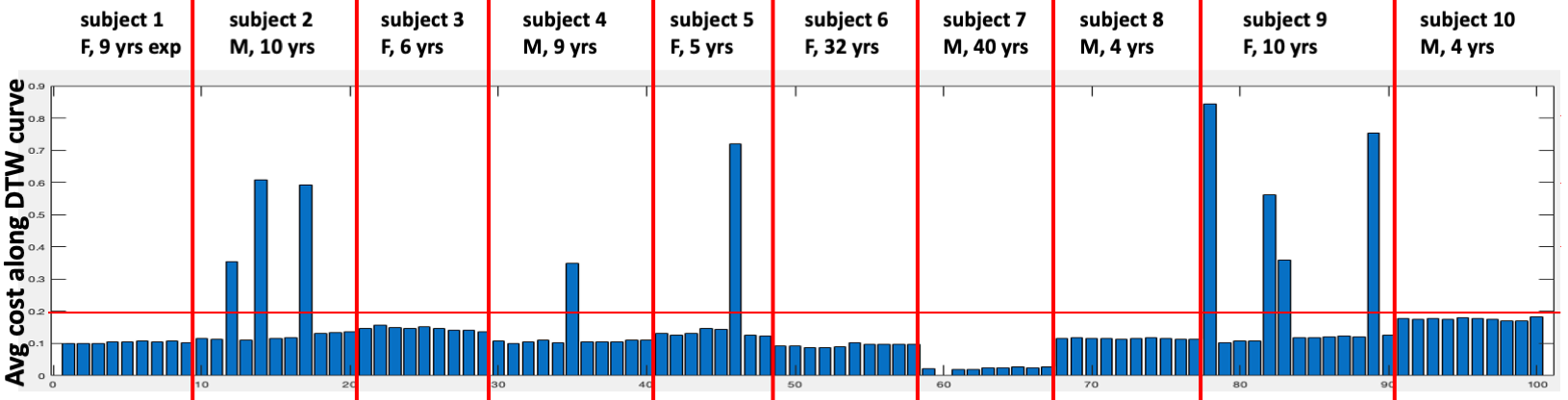}
   \caption{Alignment costs between a reference performance, subject 7 take 2, and all
   100 performances in the TMM dataset.  
    Alignment cost is computed as average cosine distance along
   the DTW curve.     
   Nine incomplete performances in the dataset are immediately identified by larger
   than average alignment costs.}
   \label{fig:100performances}
\end{figure}

To test the learned encoder, we designated one performance, subject 7 take 2, as a
special reference performance that acts as an atlas or prototype against which all other
performances are compared.  
DTW alignment on cosine distances between embedding vectors computed
by the encoder provides a
non-linear time correspondence between each performance and the reference.
\Cref{fig:100performances}  shows average cosine distance costs along the DTW
curve between all 100 performances and the reference performance.
What stands out is how the plot immediately identifies  9 partial performances 
in the dataset where a subject stopped early because of technical difficulties.  Average alignment costs between partial and complete performances are high due to the cost of accommodating
a large number of skipped frames.  Also notable are uniformly lower than usual alignment costs with other performances by  the same reference performer, subject  7, signifying consistency of performance, as well as higher than usual alignment costs with all performances of subject 10, who was the subject left out from encoder training.
\begin{figure}
  \centering 
    \includegraphics[width=0.6\linewidth]{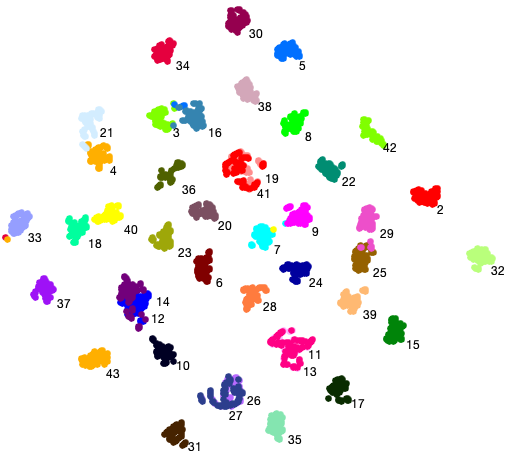}

  \caption{TSNE visualization of embedding vectors for ground truth labeled keyposes across  performances in the TMM100 dataset.   Labeled keyposes are not used when training the encoder; these compact clusters emerge from unsupervised training.}
  \label{fig:tsnevisualization}
\end{figure}

\begin{figure}[t]
  \centering
   \includegraphics[width=0.7\linewidth]{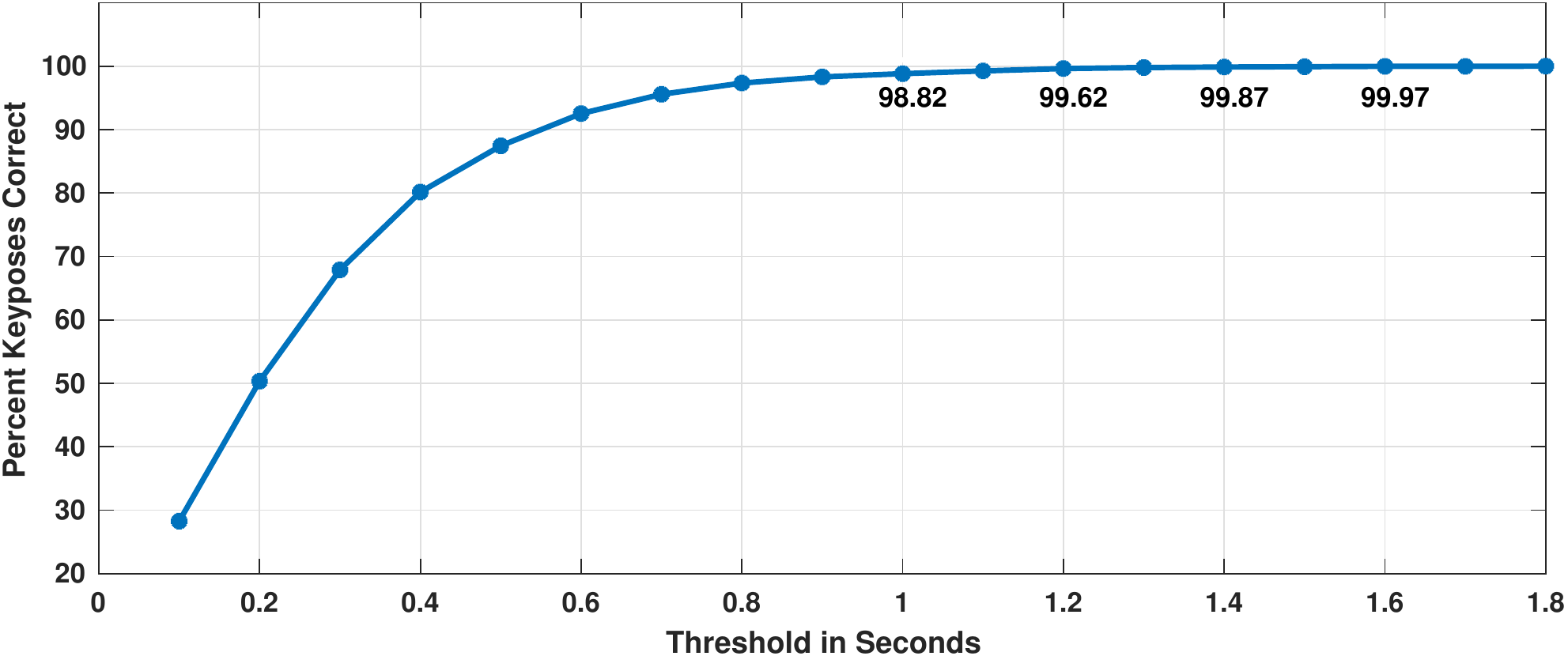}
   \caption{Percentage of transferred keypose times falling within varying thresholds of ground truth labeled times.}
   \label{fig:keyposeaccuracy}
\end{figure}

\begin{figure*}
  \centering
  \begin{subfigure}{0.47\linewidth}
    \includegraphics[width=1.0\linewidth]{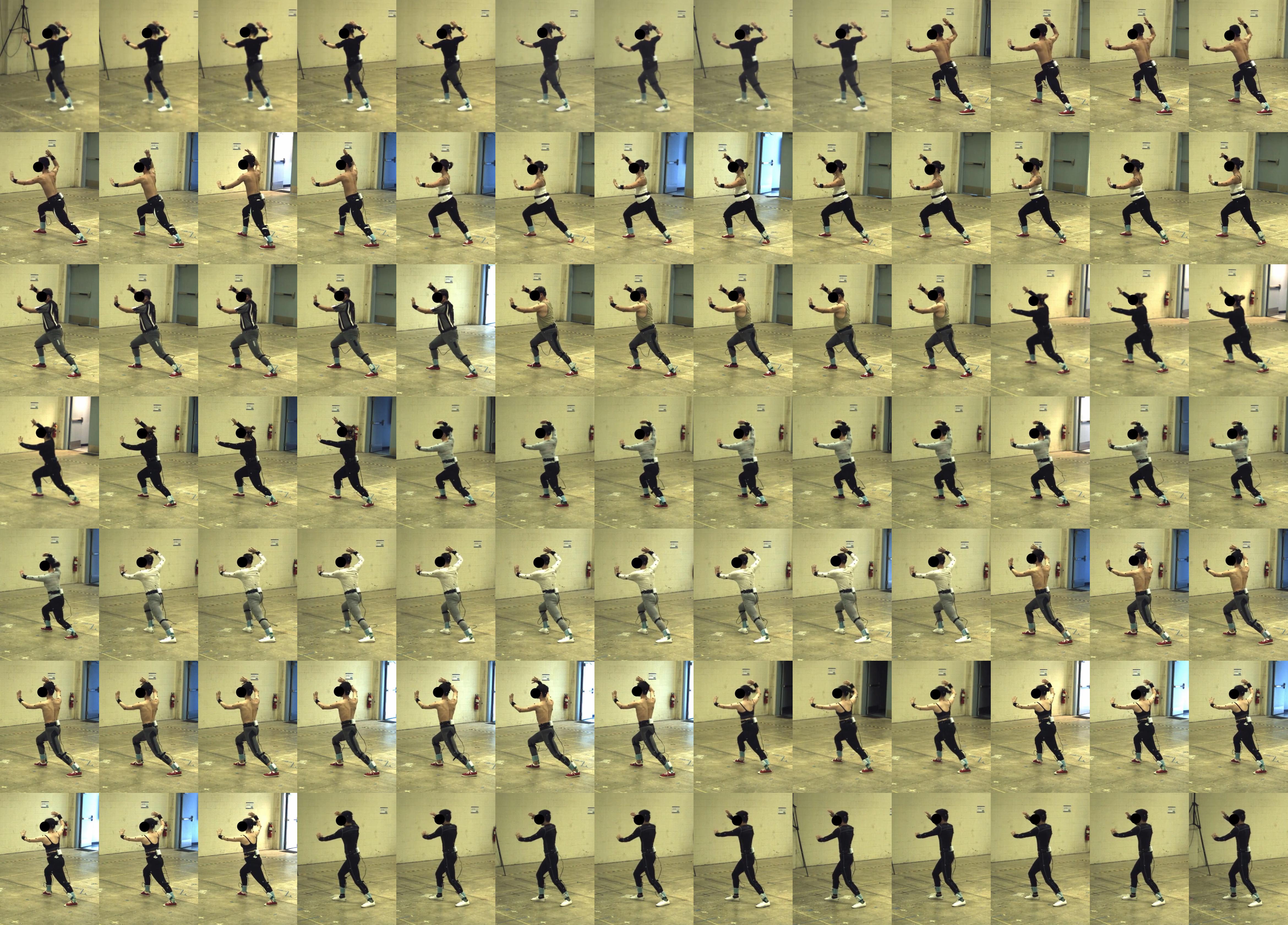}
  \end{subfigure}
  \begin{subfigure}{0.47\linewidth}
    \includegraphics[width=1.0\linewidth]{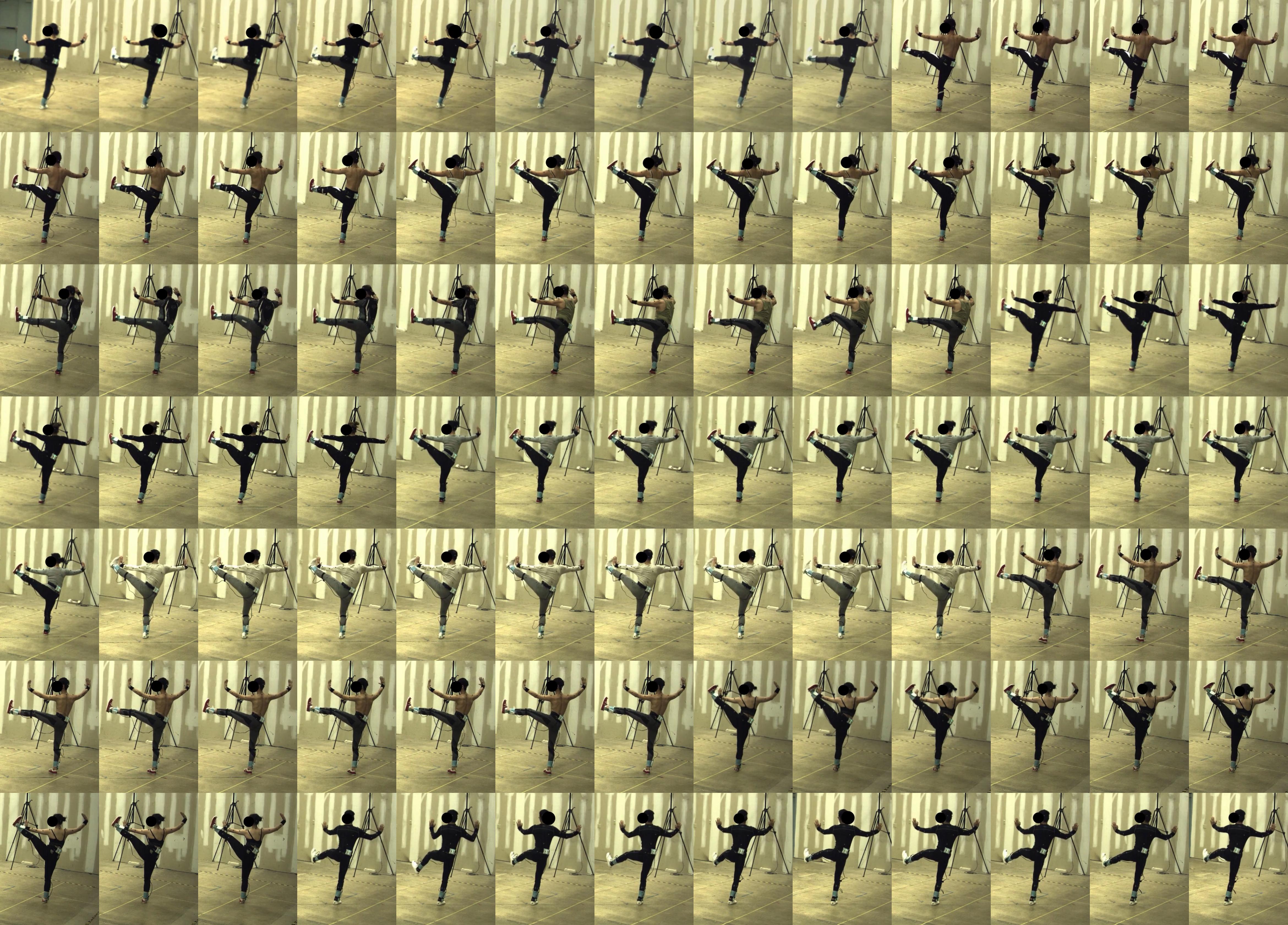}
  \end{subfigure}
    \hfill

  \caption{Sample ``contact sheet"  images showing multiple subjects performing the same pose over multiple takes, determined by temporal alignment with a reference performance (row 5 column 3) using our method for computing pose window embedding vectors followed by dynamic time warping.  }
  \label{fig:gallerypics}
\end{figure*}

For quantitative evaluation, 43 keyposes spanning the entire routine were hand labeled by Taiji practitioners in all of the performances.  Figure~\ref{fig:tsnevisualization} shows a 2D TSNE plot \cite{vanDerMaaten2008} of the 256-dimensional embedding vectors representing temporal windows centered at each labeled keypose.   Note that relatively distinct, compact clusters form for each keypose in the learned embedding space.  This property emerges in an unsupervised manner during training, suggesting a good discriminative representation has been learned.  The clusters that overlap -- 11\&13, 12\&14, 26\&27, 19\&41 -- are keyposes that label the same movements but performed at different times in the performance.  To correctly classify these keyposes requires a mechanism like DTW that uses long-range temporal context to follow the sequential order of keypose events.

To quantify alignment accuracy, labeled keypose frames in the reference performance were transferred into the other complete performances by using pairwise nonlinear DTW time alignment.  The absolute differences between DTW-predicted times and ground truth times were noted.  The results are shown in \Cref{fig:keyposeaccuracy} as a plot of percentage of predicted keyposes falling with a given time threshold  of their ground truth temporal locations. 
The predicted times are quite accurate: 87.4\% of predicted keyposes lie with 1/2 second of their GT locations; 98.9\% are within 1 second.  For a slowly performed activity like Taiji, this level of alignment accuracy provides excellent localization of matching keyposes, as can be seen qualitatively in  \Cref{fig:gallerypics} and Appendix \Cref{fig:gallerypics2}.
 Bringing multiple performances into direct correspondence
like this is useful for studying both inter and intra subject pose variability.

\subsection{UMONS-TAICHI Dataset\label{sec:umons-taichi}}

The UMONS-TAICHI dataset \cite{UMONS-TAICHI} is a 3D motion capture dataset of Taiji movements  executed by 12 participants of varying skill levels.  It is released under a CC BY-NC-SA 4.0 license at \url{https://github.com/numediart/UMONS-TAICHI}.
Each recorded performance contains a subset of 24-form Yang-style movements, 
called ``gestures" by the authors, simultaneously captured by a Qualisys mocap system (not used in this experiment) and a Microsoft Kinect V2 sensor.  They have previously analyzed a portion of the data to develop statistical algorithms for evaluating the quality of gesture performance \cite{TitsEvaluationPaper}.  

For our experiment in this section, we are interested in demonstrating automatic time alignment  of the scripted performances using the Kinect sensor data, which consists of an internal body
skeleton with 25 ``joints."   Our goal is to illustrate use of a previously learned encoder, trained on the PSU-TMM100 dataset, to align performances in a new dataset recorded by a different sensor, to demonstrate transferability of learned representations across data modalities.
To provide a comparable encoder using skeletons instead of external mocap markers
we trained a new encoder on PSU-TMM100 using its provided 17-joint Vicon data instead of the Vicon mocap marker data used previously.

\begin{table}
  \small
  \centering
  \begin{tabular}{@{}ll@{}}
    \toprule
    Vicon mocap joint & Approx Kinect joint \\
    \midrule
    \{r,l\}shoulder & shoulder\{right,left\} \\
    \{r,l\}elbow & elbow\{right,left\} \\
    \{r,l\}wrist & wrist\{right,left\} \\
    \{r,l\}hip & hip\{right,left\} \\    
    \{r,l\}knee & knee\{right,left\} \\    
    \{r,l\}ankle & ankle\{right,left\} \\            
    pelvis & spinebase \\
    waist & spinemid \\
    neck & neck \\
    clavicle & spineshoulder \\
    thorax   & spinemid \\
    \bottomrule
  \end{tabular}
  \caption{Approx mapping of Kinect joints to Vicon joints.}
  \label{tab:Kinect2Vicon}
\end{table}

\begin{figure*}
  \centering
  \begin{subfigure}{0.55\linewidth}
     \includegraphics[width=.9\linewidth]{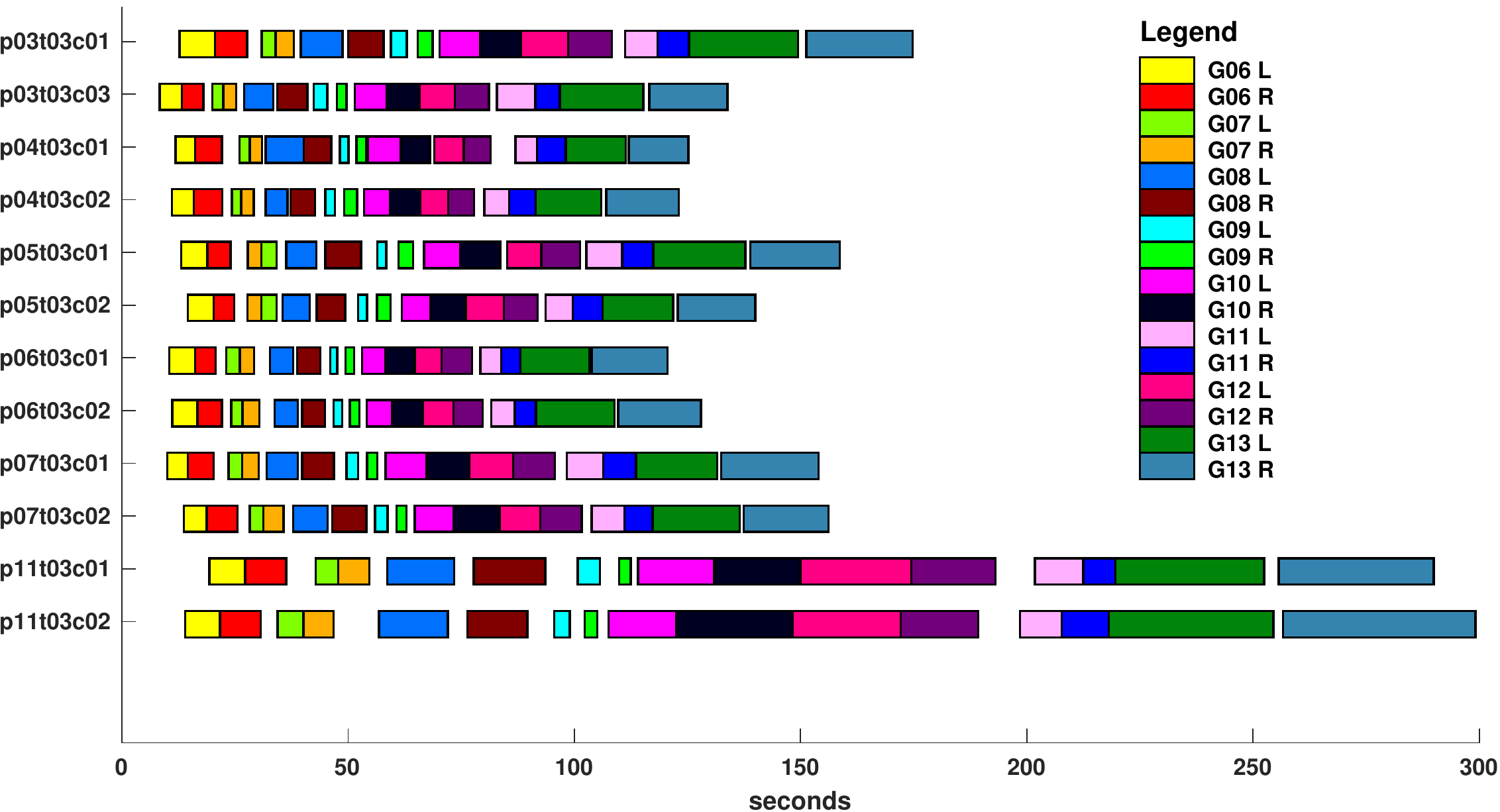}
  \end{subfigure}
  \begin{subfigure}{0.39\linewidth}
    \includegraphics[width=.9\linewidth]{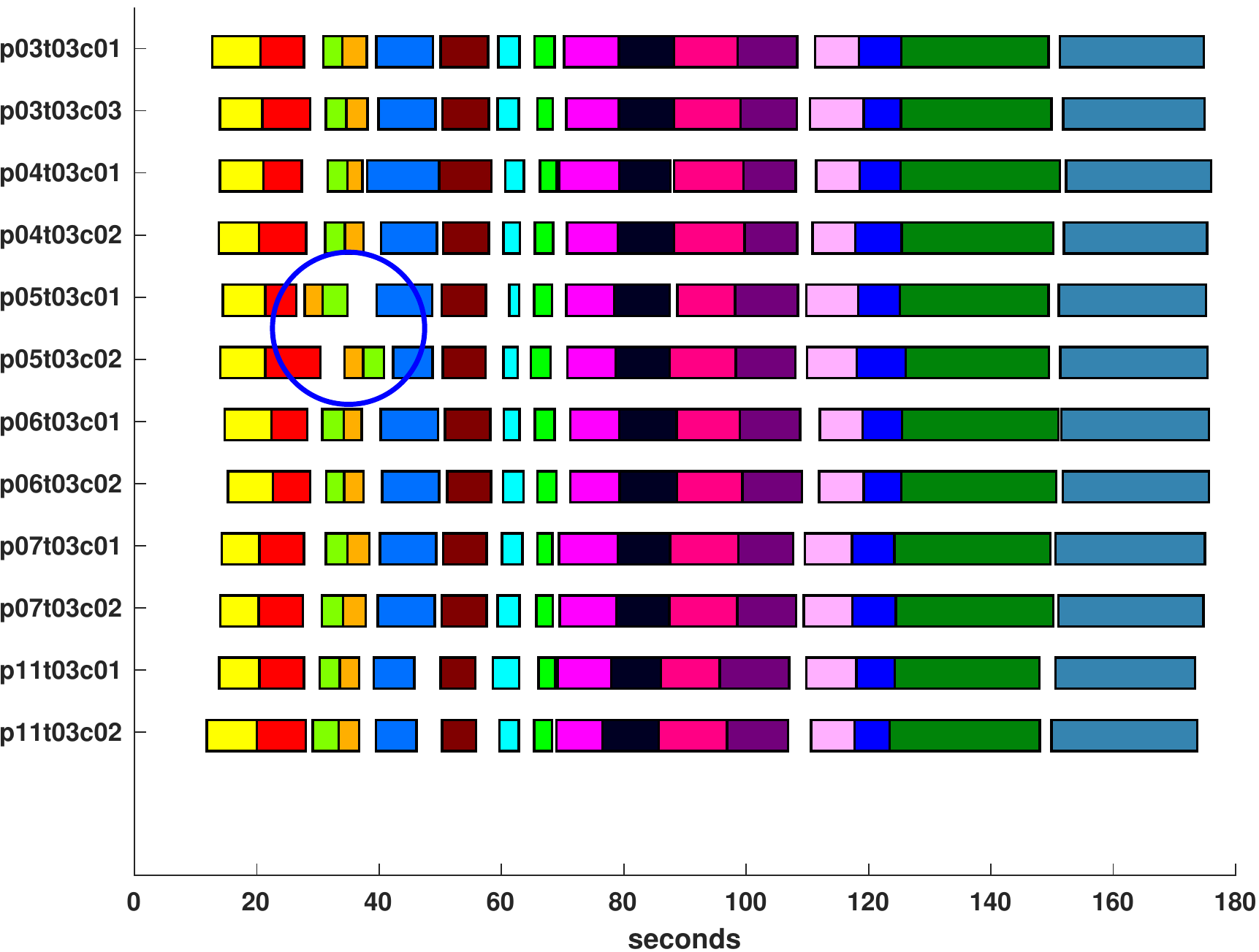}
  \end{subfigure}
    \hfill
  \caption{Manually labeled gesture start and end times provided with
  the UMONS-TAICHI dataset.  Left: raw time segments, before alignment.
  Right: segments warped into alignment with the first performance, using
  our automatically determined time alignment. Blue circle in the right
  image is explained in the text.}
  \label{fig:umonssegments}
\end{figure*}

Why is this experiment interesting?  The new dataset presents the following challenges: 1) {\bf Different sensor} - the Kinect sensor samples poses at roughly 30 fps instead of the 25 fps we used in training; it also has a non-uniform time sampling, taking longer to estimate some poses.  The 3D point trajectories are much noisier than those obtained by the Vicon system. We have made no attempt to address or correct any of these factors, preferring to see whether a learned representation can transfer without change to a new application. 
2) {\bf Different point set} - Kinect V2 provides a 25-point skeleton vs Vicon's 17-joint skeleton.  Not all Vicon joints are represented in the Kinect skeleton, and vice versa.  Joints along arms and legs have obvious counterparts but there is no direct substitute for some joints within the torso/spine, and the Kinect estimates extra points on hands, feet and head.  To ``spoof" input data that appears to be composed of Vicon joints, we make an approximate mapping as shown in \Cref{tab:Kinect2Vicon}, choosing in hard cases nearby points from the Kinect that ought to move similarly to a desired Vicon joint. Kinect {\it spinemid} is used twice, to represent Vicon {\it waist} and {\it thorax}, while 9 Kinect joints are not used at all.
3) {\bf Different routine} - although the movements in this dataset trace back to 24-form Yang-style Taiji, there are differences in how they are performed.  Despite the 24-form routine traditionally being performed while moving back and forth along a linear gait path, in this dataset the performer stays in one spot, resulting in very limited foot motion except during kicks. This stress-tests whether our encoder can recognize similarity of forms primarily from upper body movements despite being trained with both upper and lower body joints moving in tandem.  Furthermore, only a subset of the 24-form movements are performed, and transitional movements into and out of each given form are different from the training set.

We choose the T03 sessions provided in the dataset, which record 8 Taiji gestures performed in scripted order relatively continuously without pauses.  Each gesture  has a right and left handed version, for 16 activity classes in all.  We choose 12 of the T03 performances that perform L-R versions of each gesture only once per sequence (some T03 subjects perform gestures twice, L-R-L-R, before moving on to the next gesture).

 Manual segmentations of the start and end times of each gesture are provided with the dataset, and illustrated in the left side of \Cref{fig:umonssegments}. Note the difference in overall performance durations.
Choosing the first performance as a reference, and warping all other performances into alignment with it using DTW on pairwise costs between embedding vectors computed with the PSU-TMM100-trained  encoder  yields  segment alignments shown on the right of \cref{fig:umonssegments}.  There is good overlap consistency of the labeled segments for each gesture. Note that this time alignment could also map in the other direction to transfer labels from a single reference performance to identify the  temporal segments containing those labels in a new performance, providing a simple method for activity classification / localization.

The blue circle in the aligned image highlights an artifact of the dataset where one subject performed a gesture in order R-L instead of L-R.  Since the gestures occurred out of order, DTW alignment thinks one or the other is missing in that person's performance, as it can only monotonically parse gestures in the order expected in the script.  This is a limitation of DTW, not the learned encoder.  See discussion of limitations at the end of this paper.

\subsection{Karate Dataset\label{sec:karateDataset}}

The Karate Dataset \cite{karateKataDataset} contains motion capture data of two Shotokan Karate katas, {\it Bassai Dai} and {\it Heian Yondan}.  Martial arts katas are scripted exercises that are practiced to develop proper form and postures.  
The Karate kata  data was recorded at Casa Paganini InfoMus lab using a Qualisys mocap system with 25 body markers.  Seven subjects of varying experience levels were recorded performing each kata twice.  This data has been previously used to develop methods to assess quality of performance \cite{karateKataAnalyze}.
The data is available at  \url{http://www.infomus.org/karate/eyesweb_dataset_karate_eng.php}
under a CC Attribution 4.0 International (CC BY 4.0) license.

\begin{figure*}[t]
  \centering
   \includegraphics[width=0.99\linewidth]{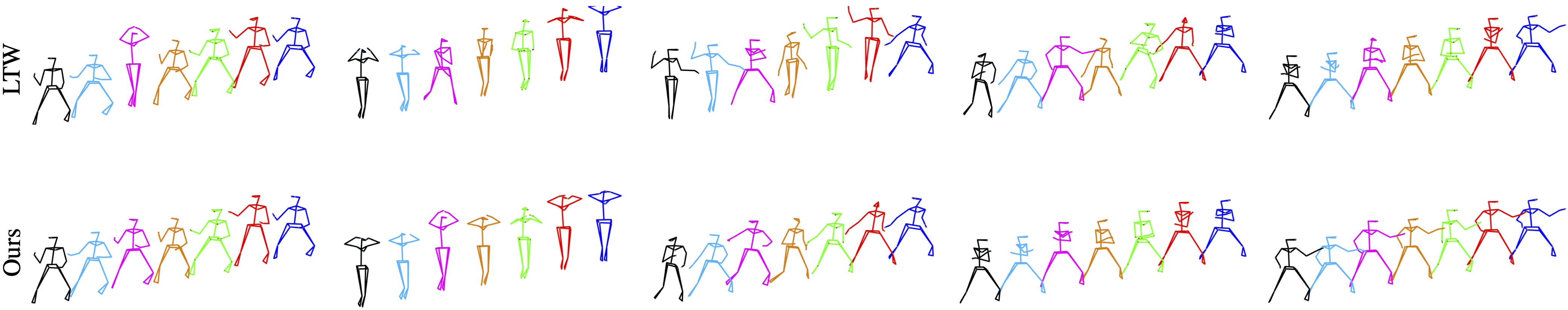}
  \caption{Top Row: Classical linear time warping (LTW) alignment does not correctly identify frames that match the pose of the blue skeleton. Bottom Row: Our alignment using DTW on cosine distances of embedding vectors, for the same sample poses.  }
  \label{fig:alignedkarateposes}
\end{figure*}

We focused on aligning seven performances of the Bassai Dai kata, one from each subject.
To again demonstrate transferability of learned representations, the  approach was to take an encoder trained previously on the PSU-TMM100 dataset, and apply it to align performances in this new dataset.
Using an encoder trained on Taiji to align Karate katas is not a crazy idea.   Although Taiji is a slow and gentle exercise, it is still a martial art, developed by abstracting many of the same types of  moves (e.g.~pushing, blocking, punching, kicking) practiced in Karate.  To apply the encoder in this new data required only changes to the procedure for generating input samples from the mocap data.  First,
an approximate correspondence mapping was specified between the 25 points measured by the Qualisys mocap system and the 39 points in the Vicon marker set that the encoder expects as input.  To save space, this mapping table is shown in the Appendix; it is constructed by hand similarly
to the Kinect-to-Vicon joint mapping in \Cref{tab:Kinect2Vicon}.  Second, we subsampled the 250 fps Qualisys data to be the 25 fps expected by the encoder, because an order of magnitude mismatch between sampling rates is not something the  encoder could be expected to overcome by itself.

No ground truth activity intervals are provided with this dataset.  However, qualitative time alignments beween the seven performances are visually good, as in the bottom row of \Cref{fig:alignedkarateposes}.   For comparison, we also show results of classical linear time warping (LTW) in the top row, using a short bow at the beginning and end of each performance as the start and end times for interpolation.  Although LTW is commonly used in biomechanics applications \cite{biomechanicsLTW}, it is more applicable for aligning short sequences like gait cycles, and less useful for longer performances or between subjects
with different experience levels, some of whom do not transition between
different forms as fluidly as others.

\subsection{SOTA Comparison on Penn Action Dataset\label{sec:PennAction}}

We use the popular Penn Action Dataset \cite{Penn_Action_Dataset} to facilitate quantitative comparison with other SOTA methods for sequence alignment.  Since 3D poses are not given with the dataset, we used Google's MediaPipe BlazePose \cite{MediaPipePose} to extract 3D body skeletons from the monocular videos.
We use the same subset of 13 non-repetitive actions and same training / validation splits as previous works\cite{Kwon_CVPR_2022,Dwibedi_2019_CVPR,Haresh_2021_CVPR}.  

Penn Action is very different from the Taiji data used above, and we made  modified variants of our algorithm to address this. 1) Because the sequences in Penn Action are very short (a few seconds),
we used temporal windows of 15 frames (.6 sec) instead of 75 (3 sec) and reduced the temporal stride of the convolution filters in the first two levels of the network to approximately compensate for the shorter temporal length.  2) Left and right pose variants of an action are considered equivalent in Penn Action  whereas our original approach respects handedness.  To allow for left-right flipping when comparing two videos A and B, the 3D pose sequence extracted from A is DTW aligned to 3D pose sequences extracted from both B and a flipped version of B, and the one with the lowest average pairwise score along the DTW   path is chosen.
3) Finally, to prevent similar poses at the start and end of a short sequence from being aligned we  introduce a linear time warping (LTW) alignment prior to play a role similar to the soft DTW prior in \cite{Haresh_2021_CVPR} or positional encoding in \cite{Kwon_CVPR_2022}, which have been shown to be crucial for good performance on this dataset \cite{Haresh_2021_CVPR}.

\begin{figure}[b]
    \includegraphics[width=.9\linewidth]{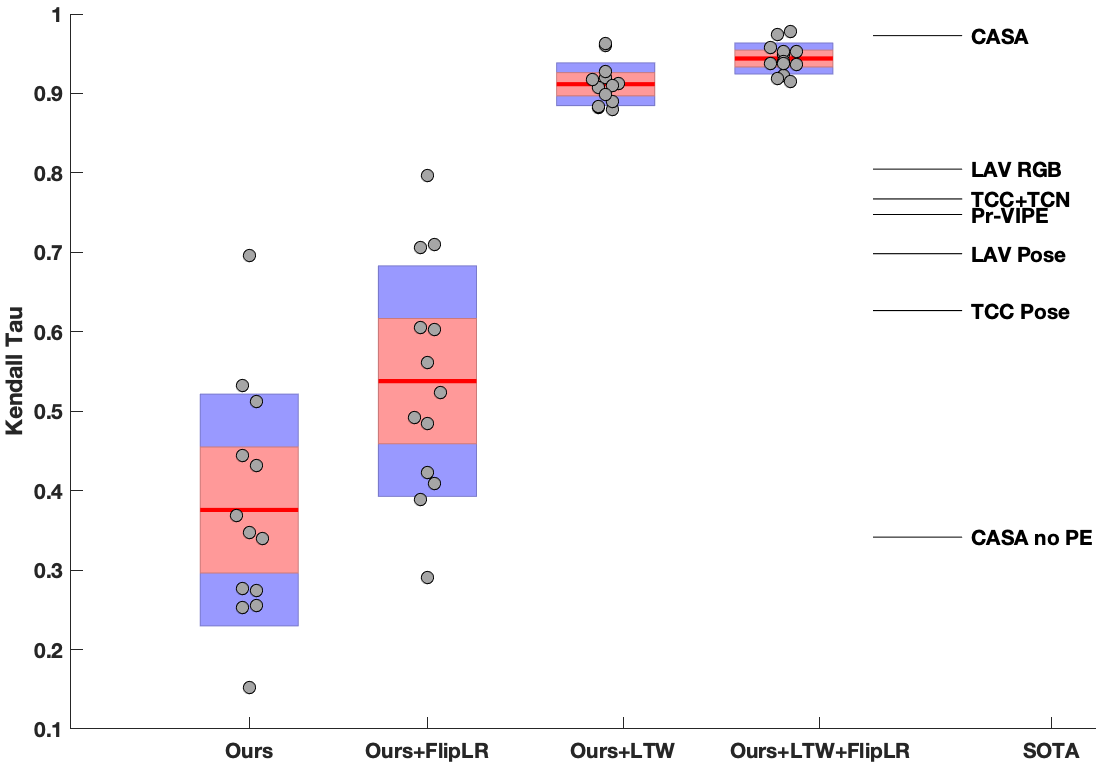}
  \caption{Kendall's Tau results on Penn Action Dataset for four variants of our approach and 
   SOTA comparison methods: CASA (with and without Positional Encoding)~\cite{Kwon_CVPR_2022}; LAV~\cite{Haresh_2021_CVPR}; TCC+TCN~\cite{Dwibedi_2019_CVPR}; Pr-VIPE~\cite{prvipe2022};
  and versions of LAV and TCC that use 3D pose data (implemented by~\cite{Kwon_CVPR_2022}).}
  \label{fig:kendalltauPA}
\end{figure}

Figure~\ref{fig:kendalltauPA} reports evaluation results of our four variants using  Kendall's Tau measure, an unsupervised measure that indicates suitability of a pairwise cost function for doing sequence-to-sequence alignment~\cite{Dwibedi_2019_CVPR}.  The first four columns are four different variants of our approach, with/without flipping and LTW.  
Each jittered data point is the average Tau value over all pairs of videos in the validation set for one  action.  The red horizontal bar is mean Tau across all 13 actions.  Also shown are 1 standard deviation around the mean (blue boxes) and standard error of the mean (red boxes).  Along the right hand side are shown reported mean Tau values for some previous SOTA methods.  See the Appendix for a tabular,  numeric version.

Notably, despite its simplicity, our full variant outperforms all previous methods except for CASA, which uses a transformer architecture with self- and cross-attention layers \cite{Kwon_CVPR_2022}.
Also notable is how much better our two variants with an LTW prior perform compared to the two that do not,  confirming the observation that a mechanism for encouraging awareness of temporal ordering and alignment coherence is needed for this dataset (evidenced also by the poor performance of CASA without Positional Encoding).

\subsection{Application to Videos in the Wild\label{sec:inthewildYoutube}}

Finally, we present qualitative results of aligning a reference mocap performance with several YouTube videos of people performing 24-Form Yang Style Taiji.  Some simple applications made possible by this alignment would be to automatically transfer captions from labeled videos to unlabeled ones,  or to generate multi-view aligned video to aid students learning the routine. It is easy to find  24-Form videos online because it is a popular routine often taught to beginners.
To extract 3D body skeletons from monocular video we again use Google's MediaPipe BlazePose \cite{MediaPipePose}, which in principle can deliver real-time performance even on mobile devices.
Although it is a state-of-art body pose detector and tracker, the 3D skeletons are more noisy than those extracted by mocap or kinect sensors due to the inherent ambiguity of extracting 3D from 2D.  This means our method has to be resilient to intermittent deficiencies such as left-right limb swaps, bodies ``tilting" towards or away from the camera in depth, and incorrect or missing locations of occluded keypoints.
Our approach can handle these types of errors because a whole temporal window of poses is mapped to each embedding vector, providing temporal context that helps smooth out intermittent errors.

As in the experiment of Section~\ref{sec:umons-taichi}, we use an encoder trained previously on Vicon joints in the PSU-TMM100 dataset, and apply it with DTW to align performances in these new videos.  This requires an approximate correspondence mapping between the 33 landmark points  estimated by BlazePose (COCO joints plus hand and face landmarks) and the 17 points in the Vicon joint set that the encoder expects as input.  This mapping table, constructed by hand, is shown in the Appendix.

A sample frame from this experiment was seen in Figure~\ref{fig:youtubepics}, and additional frames are included in  Appendix Figure~\ref{fig:moreyoutubepics}.  Here, we are aligning multiple ``in the wild" Youtube videos to a reference mocap performance recorded in a lab (upper left panel in the image). Even noisy 3D extracted pose is preferable to 2D pose, as it makes the approach insensitive to camera viewpoint. 
Because our representation of each window of poses is normalized by hip and body orientation of the center frame, our method does not rely on having a consistent 3D scene coordinate system throughout the video, making it easy to handle moving cameras.  As a result, our method  handles well a video (shown in the lower right panel of these images) that has abrupt jump cuts between camera views and an occasional narrow field of view showing upper body only.

\section{Summary and Limitations}

We have presented a simple method for unsupervised learning of an encoder that maps short 3D pose sequences into embedding vectors. 
Since the full context of poses and motion across a temporal window is encoded by each  embedding vector, simple cosine distance becomes a very discriminative pairwise cost function for which classic DTW works well to produce sequence-to-sequence alignments. When multiple scripted training sequences are available, their computed temporal alignments are mined to extract additional cross-performance matching pairs to refine the encoder. Experimental results with multiple datasets and modalities illustrate ease of use and transferability.

We finish by discussing limitations  that future work can overcome.  
The learned discriminative encodings produced in this work form a vector representation suitable for downstream tasks such as activity classification.  We have shown how sequences of poses performed in a scripted order can be aligned by dynamic time warping.  This leads naturally to a classification scheme where temporal segments of known, labeled activities are transferred through the nonlinear DTW alignments to label ranges in a new sequence, such as demonstrated on the UMONS-TAICHI dataset.  
For non-scripted activities a more general approach is needed; nonetheless, we believe our pose sequence encoding vectors form, fundamentally, a representational foundation suitable for use in more complex RNN-style classification architectures that can recognize complex activities by reasoning about sequences of primitive actions that are familiar, yet combined in a novel order not previously seen in training.  

Our data preprocessing procedure for centering and normalizing body orientation
removes information that could be useful in some applications.  For example, in a smart home application like monitoring occupant behavior in a kitchen, fixed scene locations of major appliances like refrigerator, stove, or sink form {\em functional areas} that provide valuable cues to what activities are being performed in that location.  In future work, we could explore the idea of appending positional encoding features to the input samples, either as fixed scene positions, or a more recently popularized positional encoding based on Fourier features \cite{mildenhall2020nerf,Perceiver}.

We have shown in the experiments that encoded representations learned on Vicon motion capture data can be applied to data from different sensors, e.g.~Kinect or Qualisys. 
However, to do so, we needed to specify by hand a rough correspondence between point locations in the test data format to marker or joint locations that the encoder was trained on. Can we automatically solve for these correspondences without human supervision?  One idea is to do a secondary optimization to compute an approximate linear mapping between training points and test points, either at run-time or as a secondary transfer learning stage prior to running on the test set.   
Another source of inspiration is a recent approach called SOMA that can automatically compute a robust assignment between input 3D point clouds and landmark point locations on the human body \cite{SOMA}.
\clearpage

{\small
\bibliographystyle{ieee_fullname}
\bibliography{siameseRefs}
}

\clearpage
\appendix

\section{Appendix}

\subsection{Data Normalization Description}

As discussed in Section~\ref{sec:dataNormalize} of the main paper,  temporal windows of body pose data are normalized by centering, rotating and scale normalizing the data values prior to training.  Centering and rotation use a body-specific procedure to make the 3D point data invariant to the body's location and facing direction in the capture space.  For concreteness, we describe the procedure below using Vicon Plug-in-Gait mocap markers.

\begin{figure}[b]
  \centering
   \includegraphics[width=0.7\linewidth]{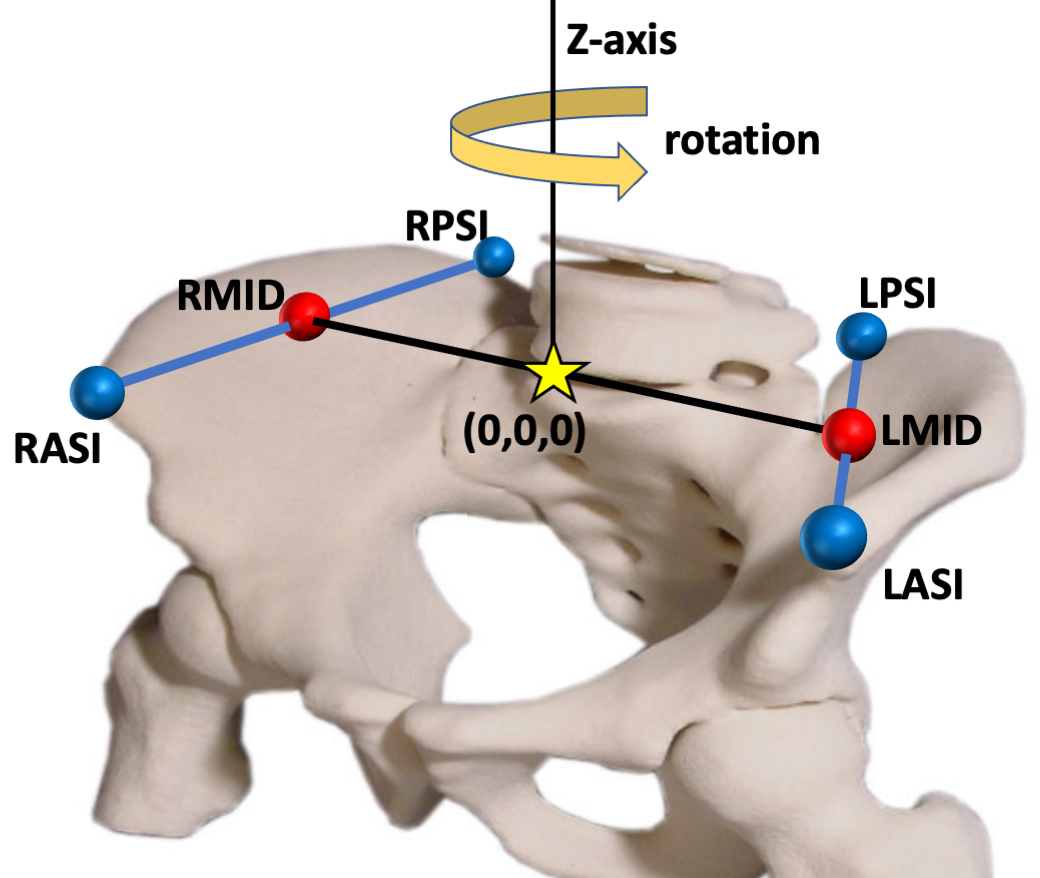}
   \caption{Domain-specific centering and rotation.}
   \label{fig:datapreparation}
\end{figure}

Consider a short-term temporal window of $n$ frames centered at time $t_0$ containing 3D trajectories of Vicon marker points. Centering of this data is performed by subtracting from each 3D point location the midpoint of the left and right anterior and posterior pelvis markers LASI$_0$, RASI$_0$, LPSI$_0$, RPSI$_0$  measured in the center frame $t_0$ of the temporal window.  Specifically, letting LMID = (LASI$_0$+LPSI$_0$)/2 and RMID = (RASI$_0$+RPSI$_0$)/2,  the origin of the 3D data in the temporal window becomes (LMID+RMID)/2.  See \Cref{fig:datapreparation}. Orientation of the body is then rotated in 2D around the mocap Z-axis (which points upward, measuring distance from the floor plane) so that the side-to-side pelvis vector RMID-LMID lies roughly along the positive mocap X-axis (specifically, the projection of RMID-LMID perpendicular to the Z-axis will lie along the X-axis) and a vector pointing forwards from the pelvis lies roughly along the positive Y-axis.  Finally, data values are normalized for scale by independently dividing each channel X, Y and Z by the standard deviation of values in that channel within the temporal window.

When only skeleton joint data is given: all variants of 3D skeleton data we have encountered include an estimated left hip and right hip joint.  Set LMID to the left hip, and RMID to the right hip, and proceed as above.

\subsection{Comparison of Loss/Cost Functions}

In this section we compare alternative loss/cost functions for aligning performances in the PSU-TMM dataset. Ultimately, what matters for sequence alignment is suitability of a learned encoding to be used to compute a pairwise distance (cost) function for doing DTW nonlinear time alignment. To evaluate  pairwise  distance functions in an unsupervised way, we use Kendall's Tau score (see \cite{Dwibedi_2019_CVPR}; also the evaluation on Penn Action Dataset).  Loosely speaking, Kendall's Tau measures suitability of a pairwise distance function for identifying corresponding frames using simple nearest neighbor classification.  Although it is not strictly valid for use on sequences like 24-form Taiji that contain repetitive movements, in this comparison we are only using it as a guide to rank-order alternatives, using sequences that all contain the same amount of repetition.

\begin{figure}[b]
  \centering
   \includegraphics[width=0.85\linewidth]{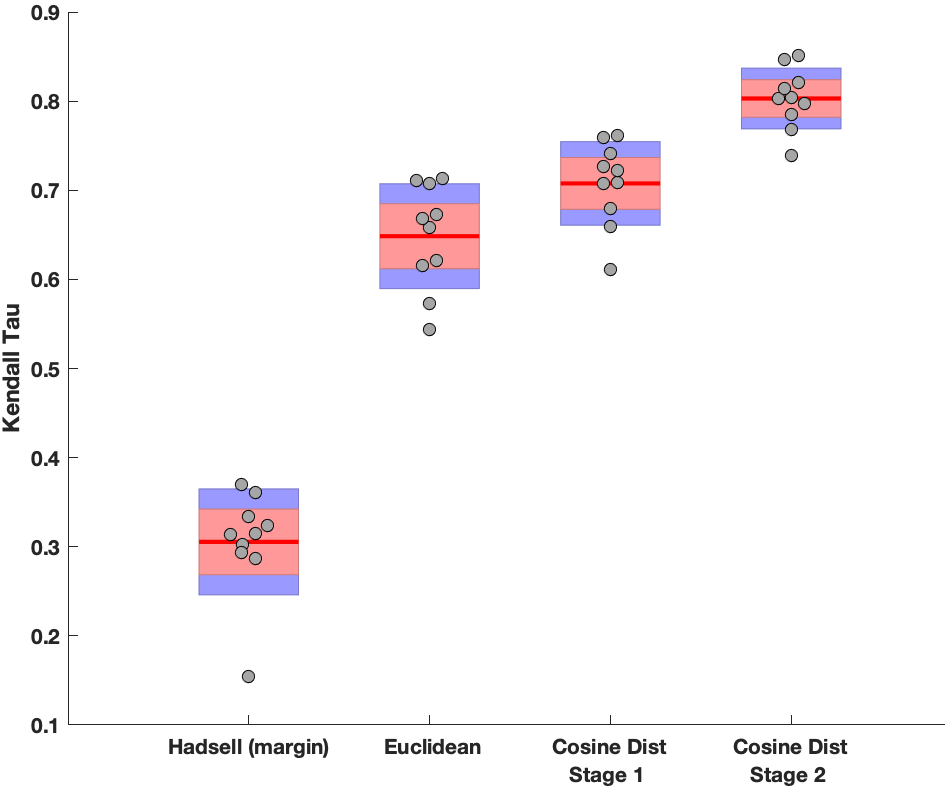} 
   \caption{Comparison of four different learned distance functions ranked left-to-right by Kendall's Tau score. See text for details.}
   \label{fig:costmatrixboxplot}
\end{figure}

\Cref{fig:costmatrixboxplot} compares four different alternatives, in increasing order of goodness from left to right.  Each box plot shows mean Kendall's Tau for 10 leave-one-subject out (LOSO) experiments on the PSU-TMM dataset (i.e. train on the performances of 9 subjects, then test alignment between all pairings of those performances with unseen performances of the 10th). Each data point is mean Kendall's Tau for one complete LOSO experiment, and each box plot summarizes those 10 data points (red horizontal line is mean value, red box shows standard error of mean, blue box depicts standard deviation).  Four alternatives are tested.
The leftmost column shows traditional max-margin contrastive loss, due to Hadsell~\cite{HadsellEarlySiamese}.  This promotes positive pairs to map close together in embedding space, and negative pairs to be pushed further apart than a margin distance.  After learning an embedding, the distance function for DTW is Euclidean distance in embedding space.
The other 3 alternatives are based on using our cosine similarity loss function (see main paper) for training an embedding space where positive pairs map to vectors with cosine similarity 1 and negative pairs map to vectors with cosine similarity 0.
Column 2 evaluates using Euclidean distance between those learned embedding vectors as the pairwise distance function.  Intuitively this makes sense because, as seen with the TNSE visualization of keyposes,  similar poses+motions map to  clusters that are  compact as well as distinct from each other.
Columns 3 and 4 are the results of our proposed approach after Stage 1 and Stage 2 training.   That is, cosine similarity loss is used for training an embedding space, and then cosine distance is used as the  pairwise distance function.  Stage 1 uses only within-performance data (augmented to form positive pairs), while Stage 2 harvests additional cross-peformance positive pairs based on DTW alignment using the initial pairwise distance functions from Stage 1.

\begin{figure}[t]
  \centering
  \begin{tabular}{c c}
   \includegraphics[width=0.45\linewidth]{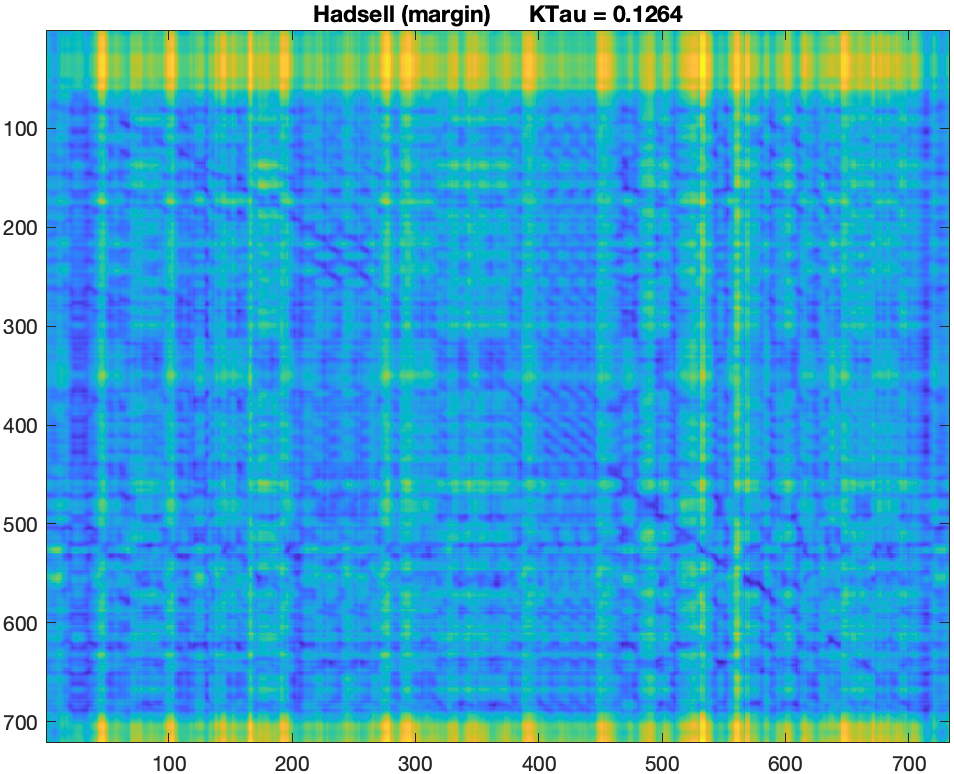} &
   \includegraphics[width=0.45\linewidth]{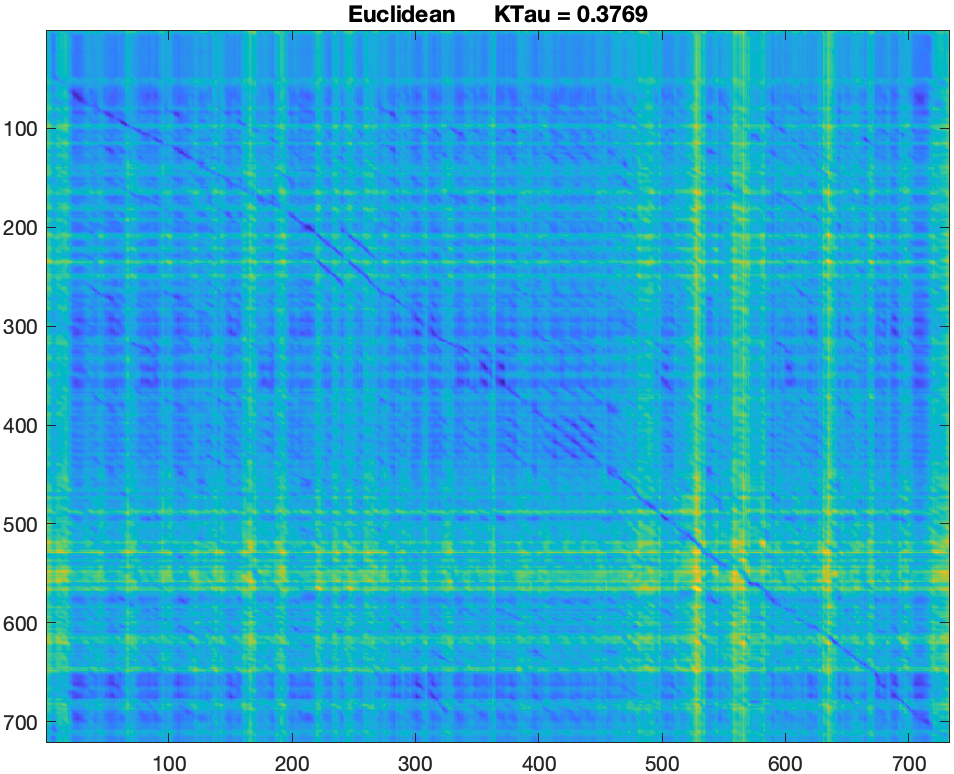} \\
    \includegraphics[width=0.45\linewidth]{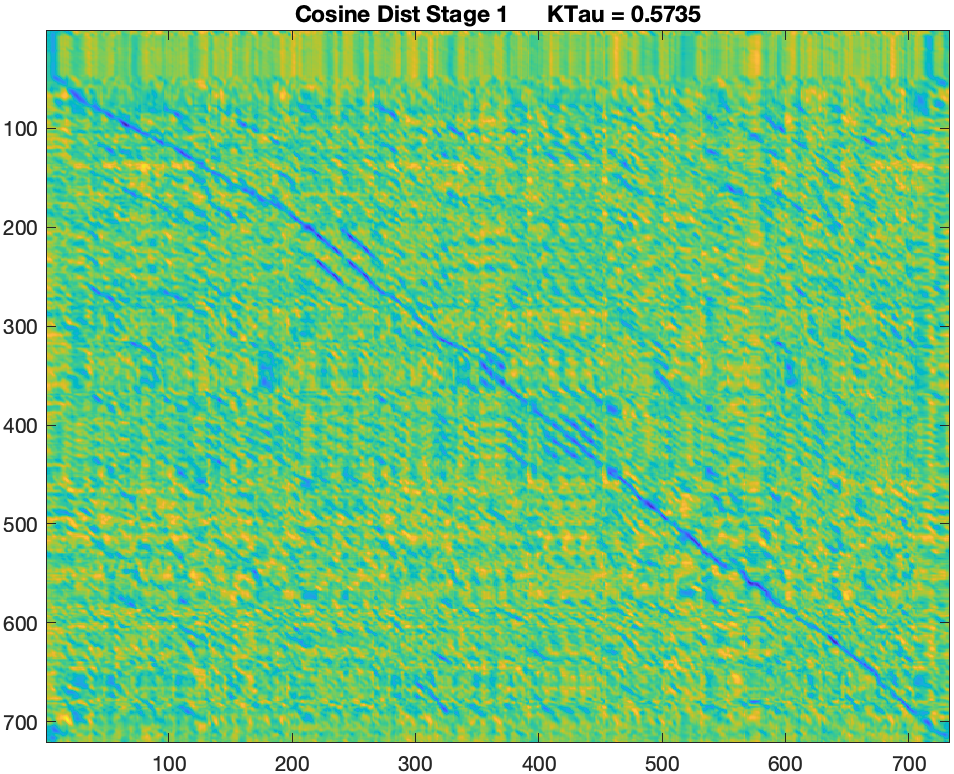} &
   \includegraphics[width=0.45\linewidth]{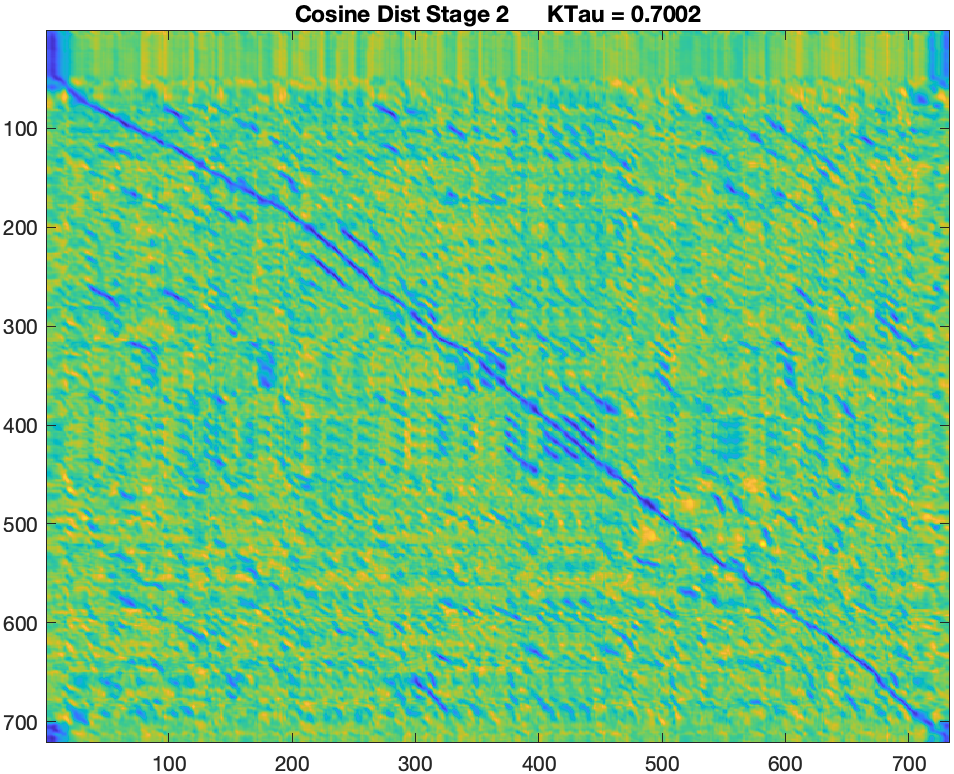} 
   \end{tabular}
   \caption{Visual comparison of four different pairwise distance functions and their associated Kendall's Tau score, for the same pair of performances to be aligned.  Our approach after Stage 2 training (lower right) yields the most well-defined pairwise distance matrix for use by DTW alignment.}
   \label{fig:costmatrixtaus}
\end{figure}

The ranking provided by these summary results is also borne out for specific pairs of performances to be aligned.
\Cref{fig:costmatrixtaus} visualizes the pairwise distance functions for these four alternatives between two performances (subject 1 take 1 and subject 5 take 1).  What we are looking for visually in a distance function for alignment is a clear, high-contrast monotonic path running from from upper left to lower right.

\subsection{Numeric Data for Penn Action Evaluation}

The evaluation on Penn Action Dataset in terms of Kendall's Tau score, which was visualized graphically in Figure~\ref{fig:kendalltauPA} of the paper, is based on the following numeric table (\Cref{tab:KendallTau}).  Each Tau score is the reported mean Tau computed over 13 non-repetitive actions in the Penn Action dataset.  
Aside from results for the four variants labeled ``ours", all other reported Tau values are quoted from Table 2 of \cite{Kwon_CVPR_2022}, except for TCC+TCN which is from Table 6 of ~\cite{Dwibedi_2019_CVPR}.
Despite its simplicity, a variant of our approach allowing left-right flips and a linear time warping prior outperforms all previous methods except for CASA, which uses a transformer architecture with self- and cross-attention layers \cite{Kwon_CVPR_2022}.

\begin{table}[h]
  \small
  \centering
  \begin{tabular}{@{}ll@{}}
    \toprule
    Method   & Mean $\tau$  \\
    \midrule
CASA & {\bf 0.9728} \\
ours+FlipLR+LTW & \underline{\bf0.9440} \\
ours+LTW & 0.9116\\
LAV RGB & 0.8047 \\
TCC+TCN & 0.7672 \\
Pr-VIPE & 0.7476 \\
TCC RGB & 0.7012 \\
LAV Pose & 0.6983 \\
SAL & 0.6336 \\
TCC Pose & 0.6267 \\
ours+FlipLR & 0.5379\\
ours (basic) & 0.3757\\
CASA w/o PE & 0.3415 \\
    \bottomrule
  \end{tabular}
  \caption{Kendall's Tau results on Penn Action Dataset for
 four variants of our approach and SOTA comparison methods.
 Best result (highest Tau score) is in bold, second best is bold underlined.}
  \label{tab:KendallTau}
\end{table}

\subsection{Qualisys to Vicon Marker Set Mapping}

The Karate dataset experiment in Section~\ref{sec:karateDataset} aligned Karate kata data recorded with a Qualisys mocap system having 25 markers \cite{karateKataDataset}, using an encoder trained on Taiji data recorded with a Vicon mocap system having 39 markers \cite{PlugInGaitReference}.  To map the 25-marker data into the 39 markers that the encoder expects to see, we used the approximate mapping table shown in \Cref{tab:Qualisys2Vicon}.  Note that several Qualisys markers ``double up" to cover two different Vicon markers -- for example a single back of head Qualisys marker does double duty as both right and left back of head Vicon markers.  As discussed in the main paper,  when directly corresponding points are not available, substituting a nearby point that moves similarly works fine.

\begin{table}[h]
  \footnotesize
  \centering
  \begin{tabular}{| l l | l l | l l |}  
    \hline
    Vicon &  Qualisys & Vicon &  Qualisys & Vicon &  Qualisys \\
    \hline
    LFHD & LFHD &     LSHO & LSHD &     LTHI &  LFHP \\
    RFHD & RFHD &    LUPA & LSHD &    LKNE & LKNE \\ 
    LBHD & BKHD &    LELB & LELB &   LTIB & LKNE \\
    RBHD & BKHD &    LFRM & LELB &     LANK & LBAK \\
    C7 &  C7 &   LWRA & LWRS &    LHEE & LBAK \\
    T10 & C7 &   LWRB & LWRS &     LTOE & LFAK \\
    CLAV & NCK &    LFIN & LIND &    RTHI & RFHP \\
    STRN & NCK &    RSHO &  RSHD &    RKNE & RKNE \\
    RBAK & RSHD &    RUPA & RSHD &    RTIB & RKNE \\
    LASI & LFHP &    RELB & RELB &    RANK & RBAK \\
    RASI & RFHP &  RFRM &RELB &     RHEE & RBAK \\
    LPSI & LBHP &    RWRA & RWRS &    RTOE & RFAK \\
    RPSI & RBHP &     RWRB & RWRS &  \  & \ \\
    \ & \                 &   RFIN & RIND & \  &  \ \\
    \hline
  \end{tabular}
  \caption{Approx mapping of  25 Qualisys marker points used in \cite{karateKataDataset} to
  the 39 Vicon Plug-in-Gait markers \cite{PlugInGaitReference}.}
  \label{tab:Qualisys2Vicon}
\end{table}

\subsection{BlazePose to Vicon Marker Set Mapping}

The in-the-wild Youtube video experiment and the Penn Action Dataset evaluation make use of Google's MediaPipe BlazePose \cite{MediaPipePose} software to extract 3D skeletons from monocular 2D videos.  BlazePose estimates 33 landmark points on body, face and hands.  To process this data using an encoder trained previously on the Vicon Plug-in-Gait 17-joint skeleton point set, we use the mapping shown in Table~\ref{tab:Blaze2Vicon}.

\subsection{Additional Aligned Examples}

\begin{figure*}[p]
  \centering
  \setlength{\tabcolsep}{1pt}
  \begin{tabular}{| c | c |}
  \hline
 \includegraphics[width=.48\textwidth]{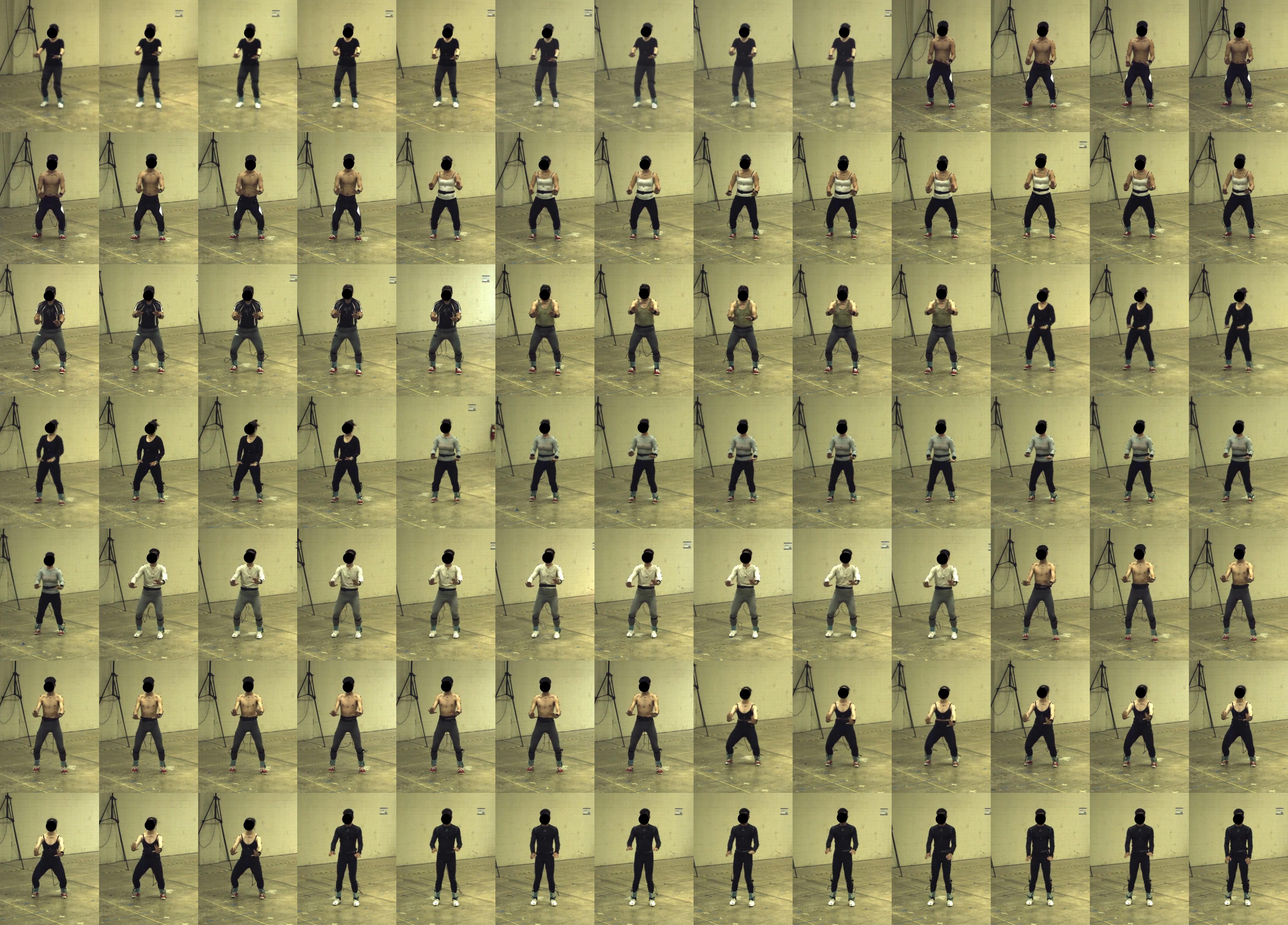} &
    \includegraphics[width=.48\textwidth]{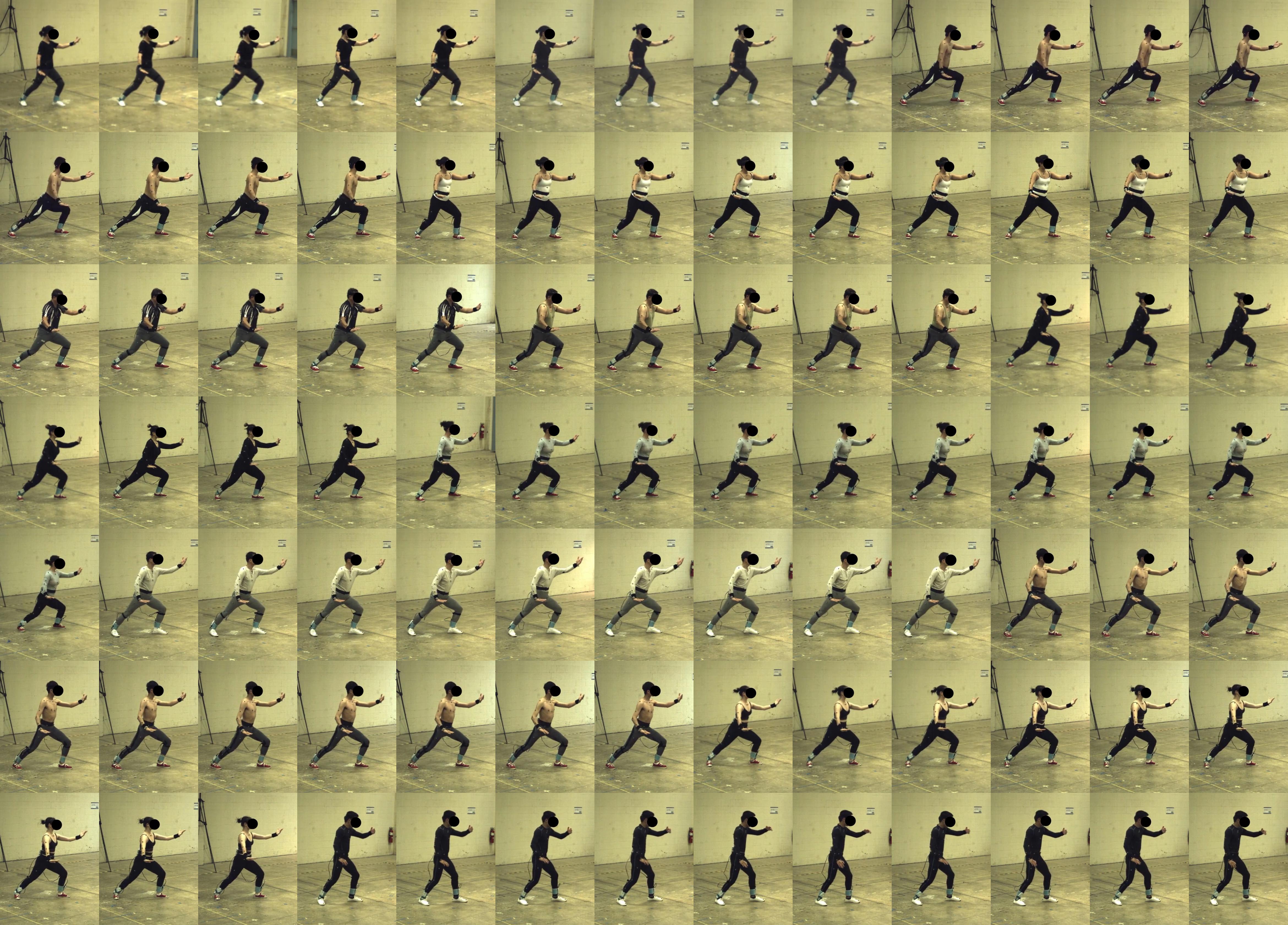} \\   \hline
 \includegraphics[width=.48\textwidth]{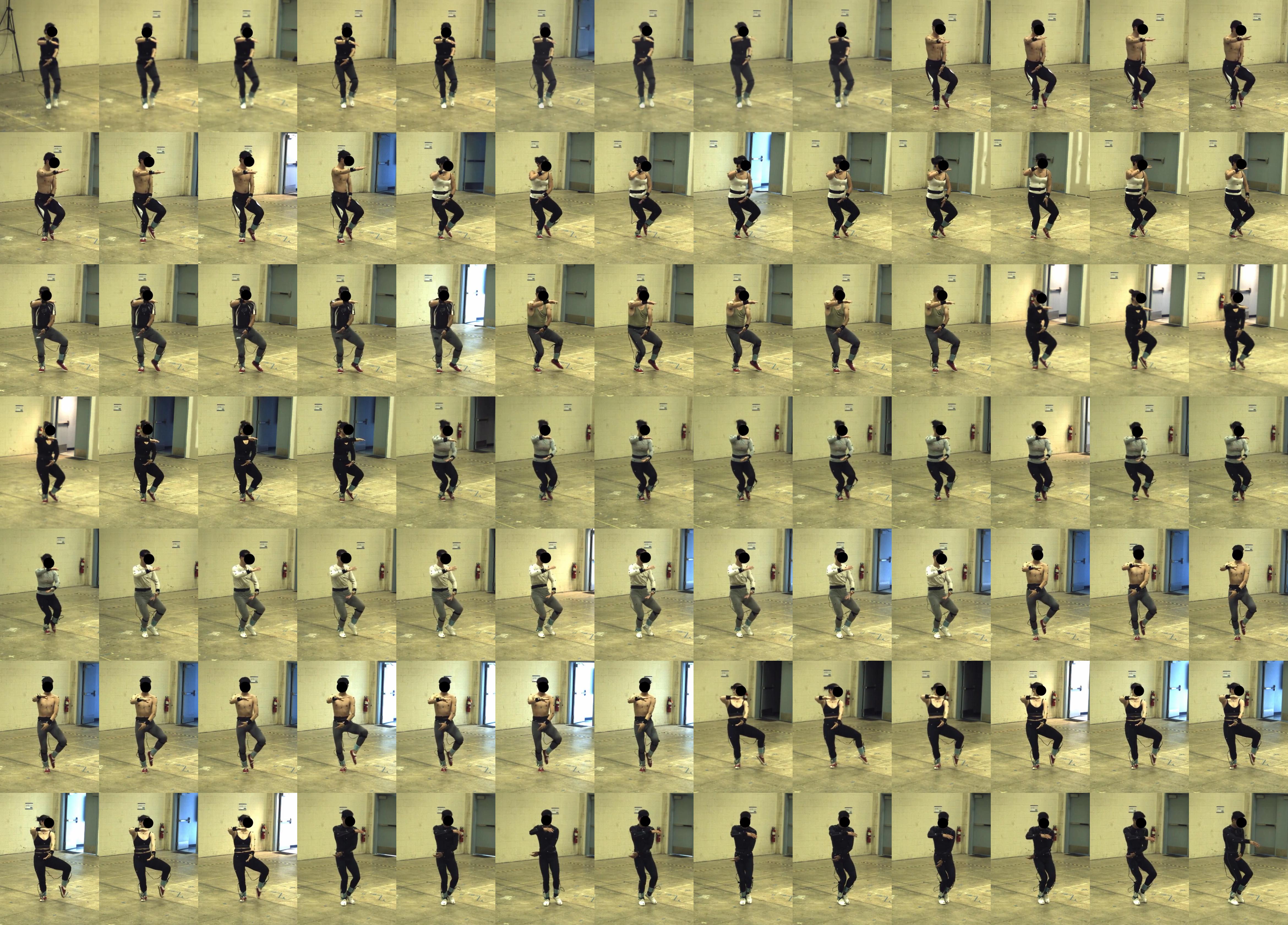} &
    \includegraphics[width=.48\textwidth]{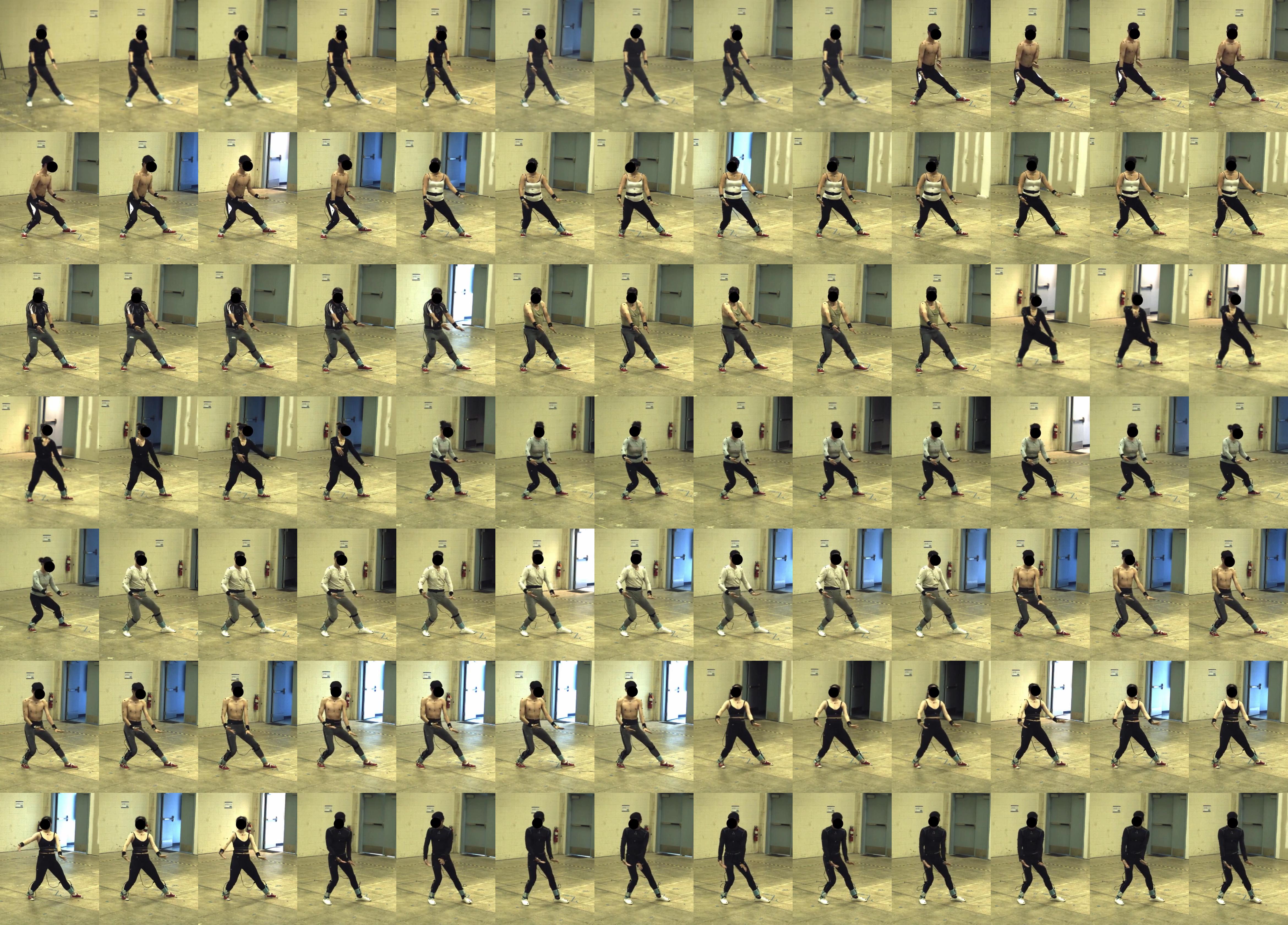} \\  \hline
 \includegraphics[width=.48\textwidth]{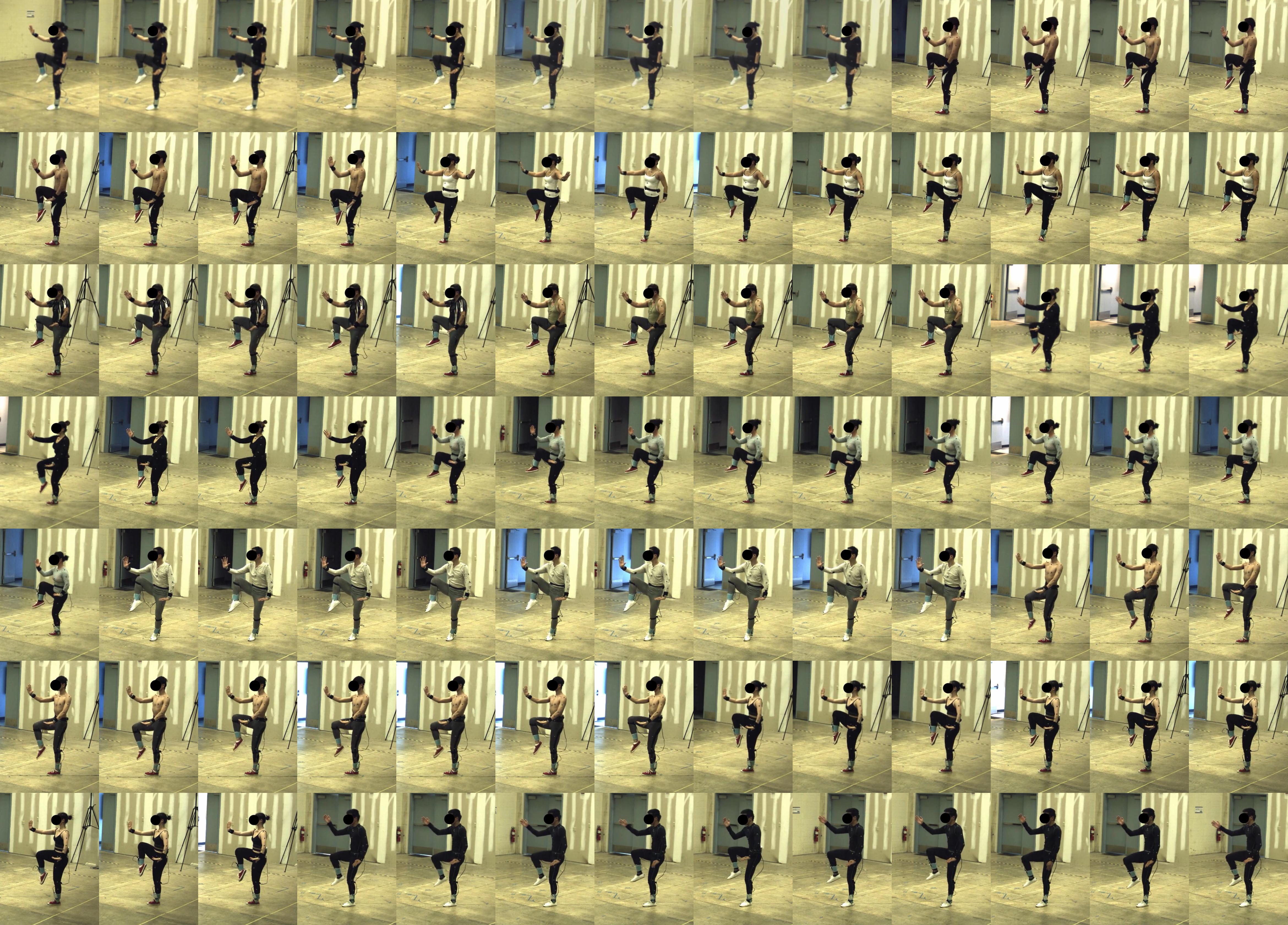} &
    \includegraphics[width=.48\textwidth]{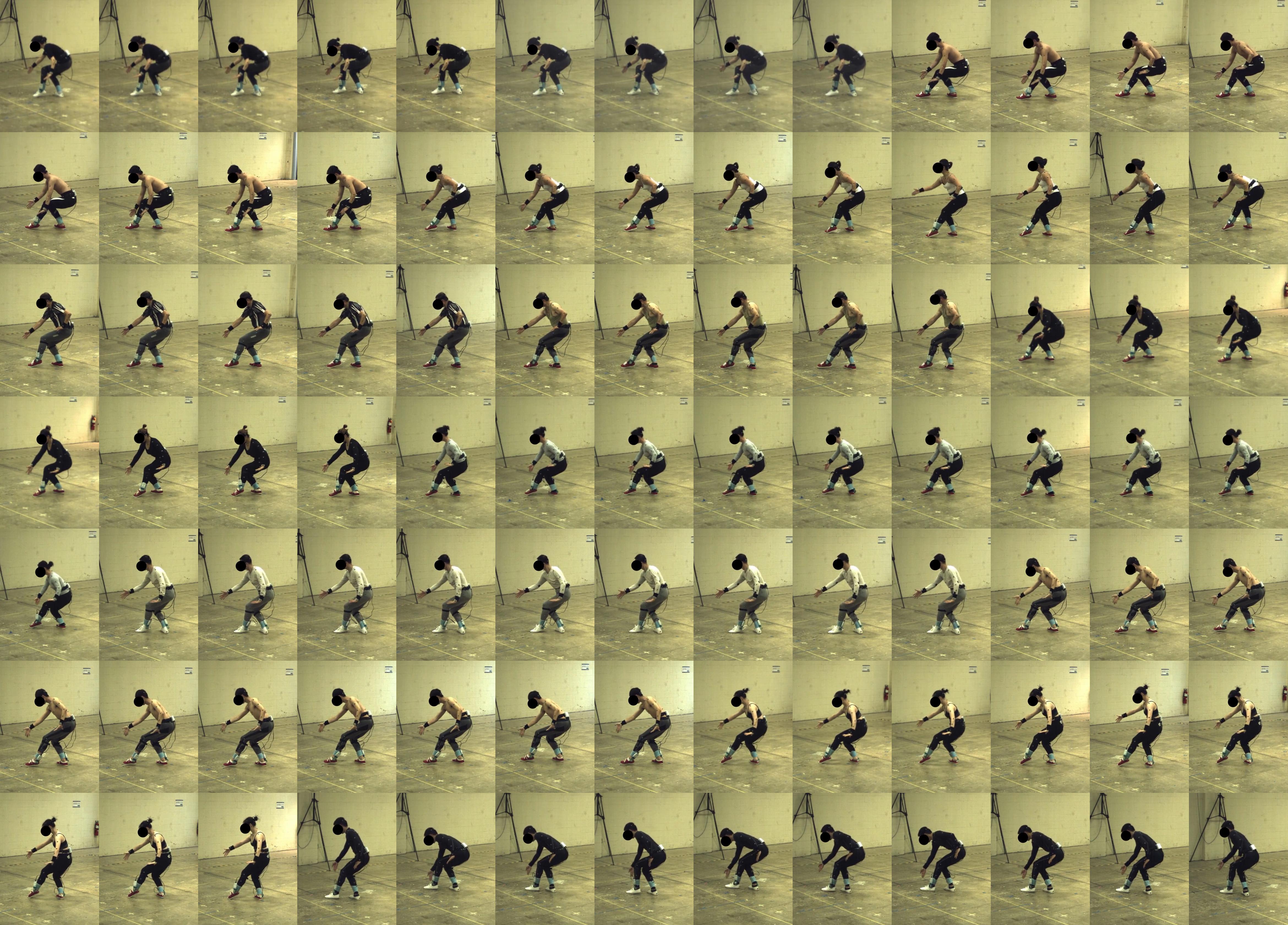} \\            \hline
      \end{tabular}
    \vspace{-1.5ex}
  \caption{Six additional ``contact sheet"  images illustrating snapshots of time alignment results for 91 performances by 10 different subjects.  The reference performance  that all others have been aligned to appears in row 5 column 3 of each image array.
  }
  \label{fig:gallerypics2}
\end{figure*}

Section~\ref{sec:tmm100dataset} Figure~\ref{fig:gallerypics} presented results of aligning multiple 24-form Taiji performances with a single reference performance.   \Cref{fig:gallerypics2} shows additional ``contact sheet" images depicting time-aligned poses for 91 complete  performances performed by 10 different subjects in the PSU TMM Dataset.  These are achieved by aligning each performance to one reference performance used as an atlas or prototype.  The reference performance, subject 7 take 2, appears in row 5 column 3 of each of these image arrays.

\begin{figure*}[p]
  \centering
  \setlength{\tabcolsep}{1pt}
  \begin{tabular}{ c  c  c }
 \includegraphics[width=.48\textwidth]{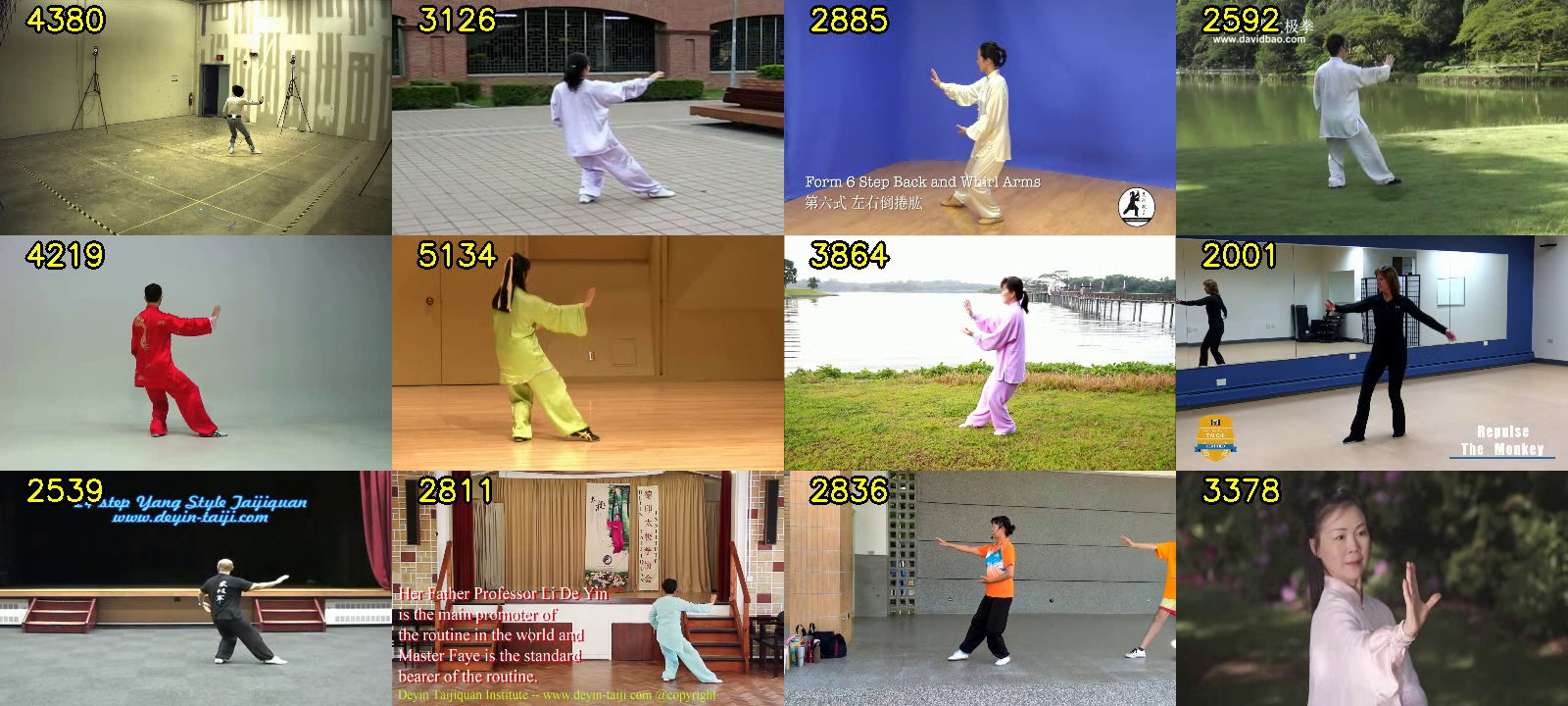} & \ \ \ \ \ &
    \includegraphics[width=.48\textwidth]{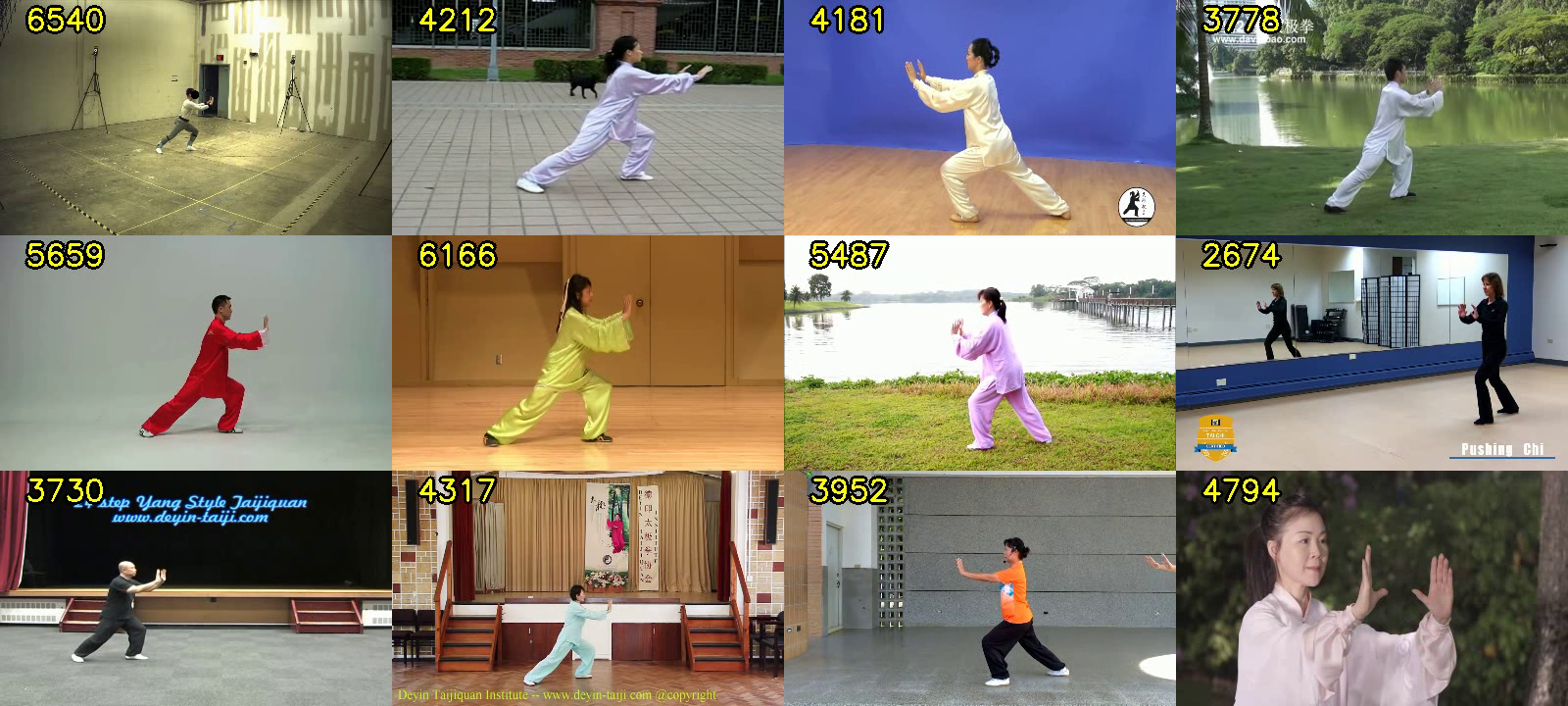} \\   
    \\
 \includegraphics[width=.48\textwidth]{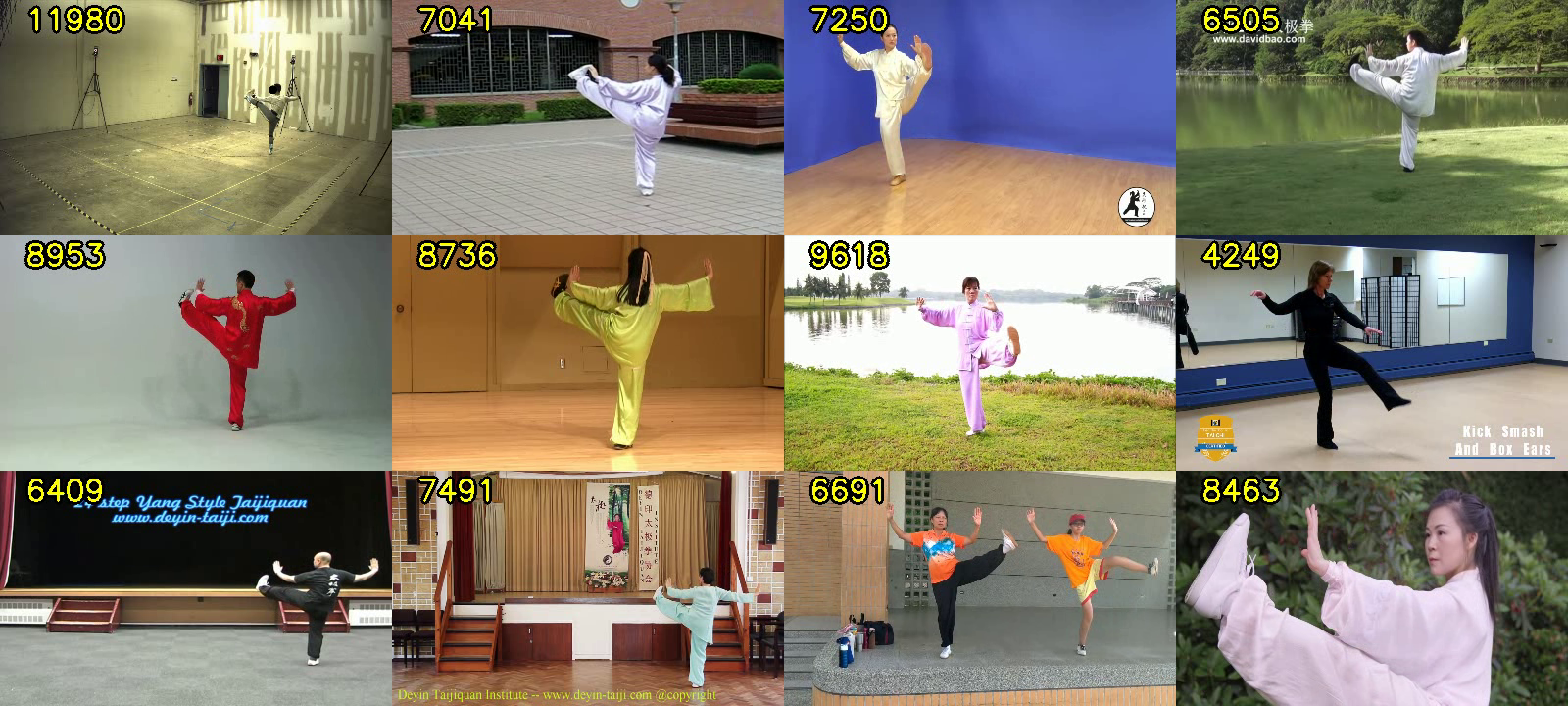} & &
    \includegraphics[width=.48\textwidth]{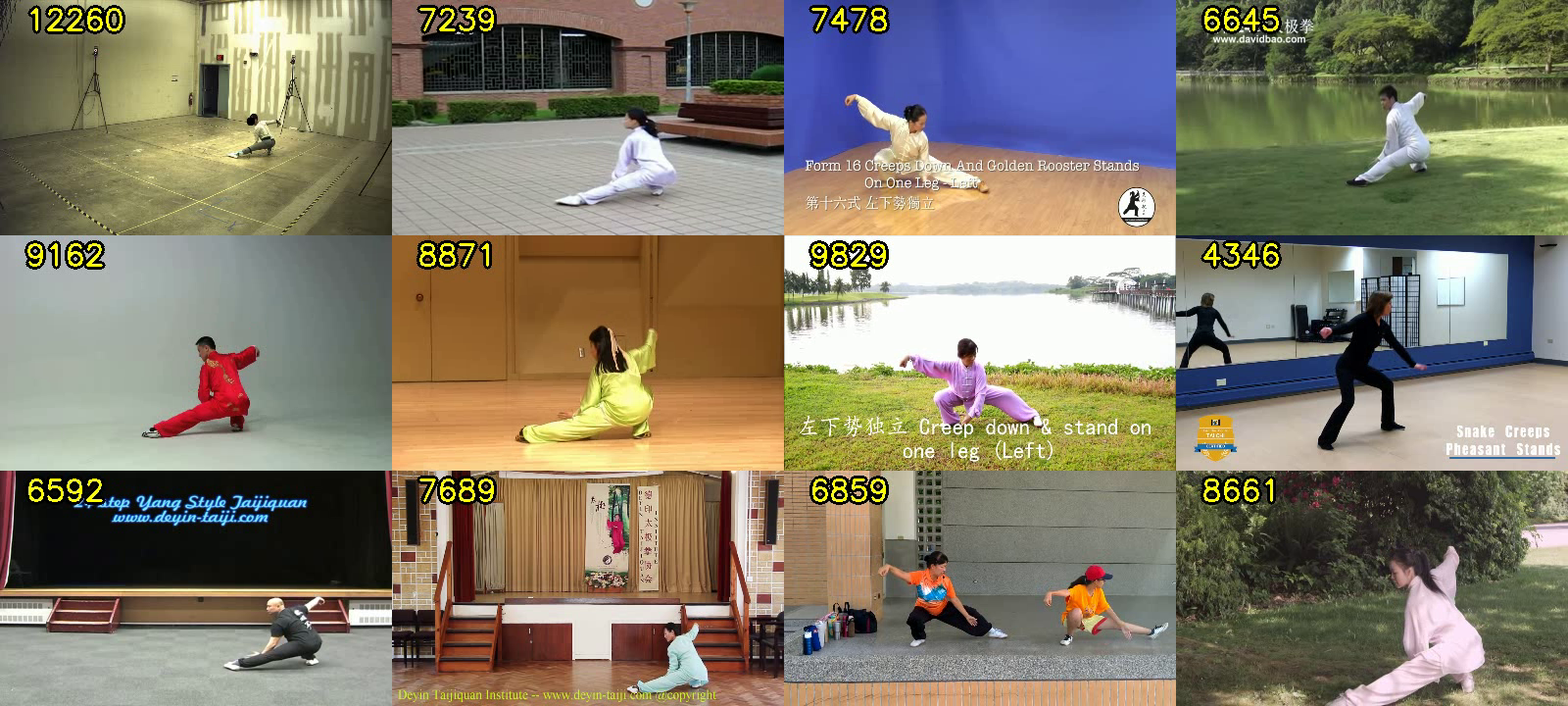} \\  
    \\
  \includegraphics[width=.48\textwidth]{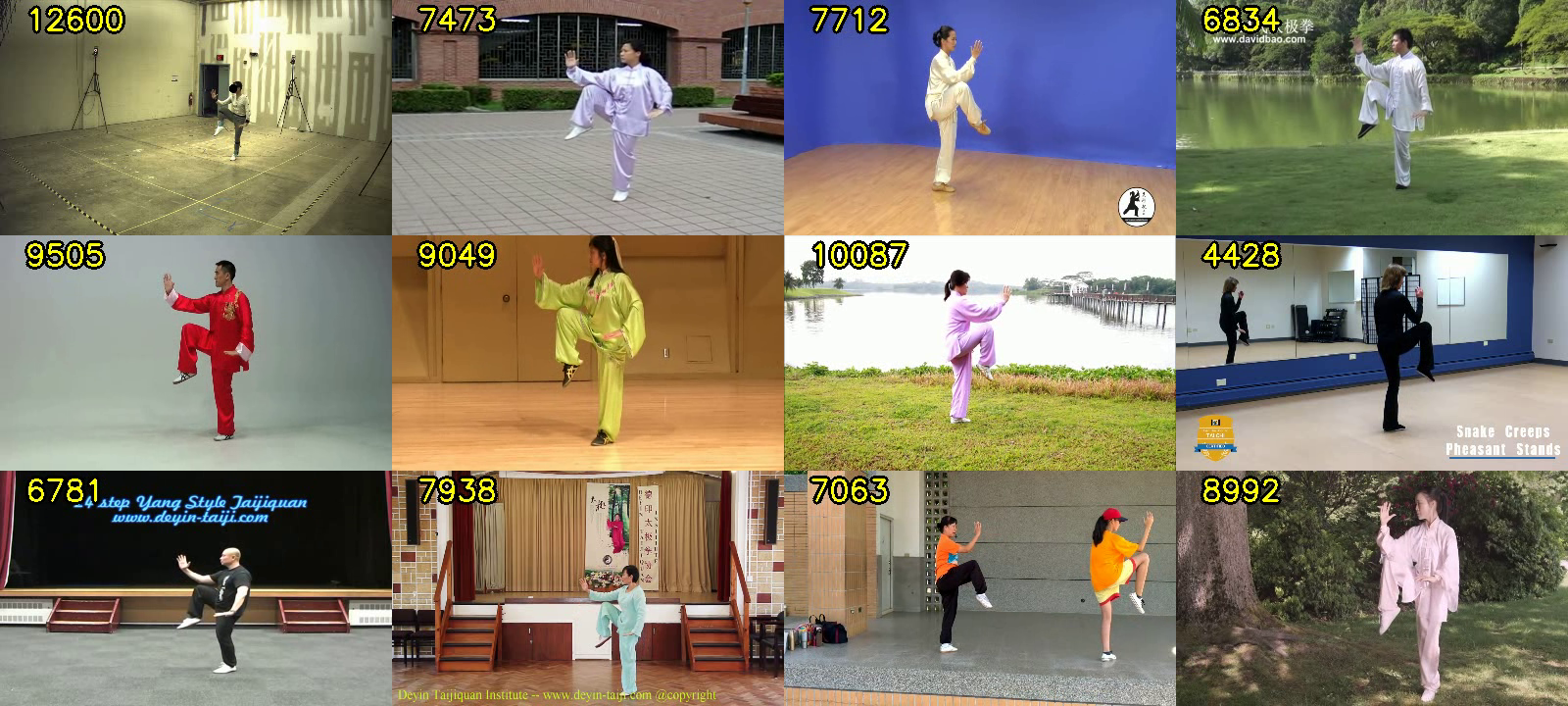} & &
    \includegraphics[width=.48\textwidth]{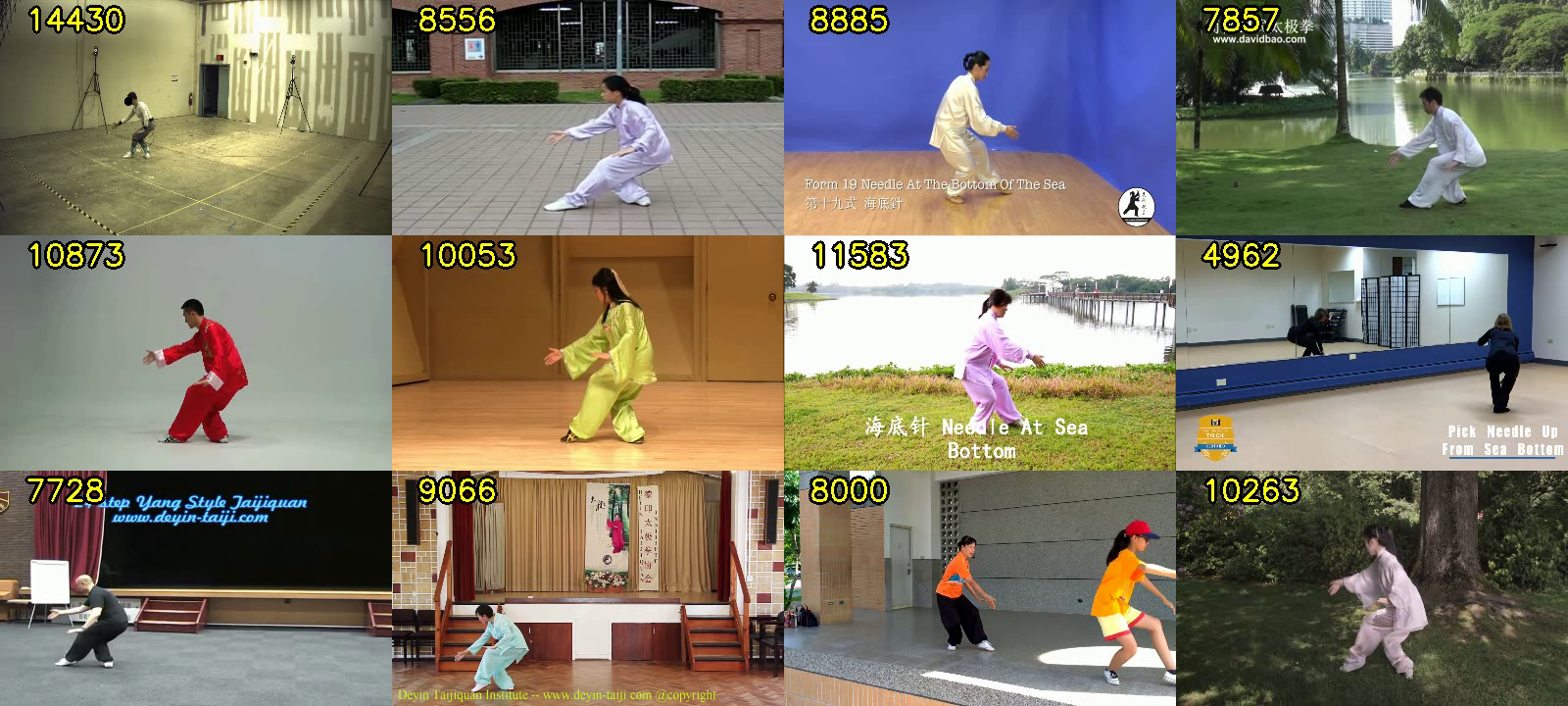} \\  
      \end{tabular}
    \vspace{-1.5ex}
  \caption{Additional examples of the alignment of multiple ``in the wild" Youtube videos to a reference  performance captured in a mocap lab (upper left panel).  These are still frames from a continuous video showing alignments for the complete 24-form Yang-style Taiji routine.
  }
  \label{fig:moreyoutubepics}
\end{figure*}

Section~\ref{sec:inthewildYoutube} addressed ``in the wild" monocular 2D video recordings from Youtube, and \Cref{fig:youtubepics} showed results of aligning a reference performance recorded in a Mocap lab (upper left panel) with Youtube videos of people performing the same 24-form Taiji routine.  Additional examples of this alignment captured at different times/poses of the reference performance are shown in Figure~\ref{fig:moreyoutubepics}.  It is particularly notable that, without any additional modification, our method  handles well the video in the lower right corner, which was shot artistically with jump cuts  between camera views and occasional tight zooms showing only the upper body.
\newpage

\begin{table}[ht]
  \small
  \centering
  \begin{tabular}{@{}ll@{}}
    \toprule
    Vicon Joint & Approx BlazePose Landmark \\
    \midrule
    \{r,l\}shoulder & \{right,left\}\_shoulder \\
    \{r,l\}elbow & \{right,left\}\_elbow \\
    \{r,l\}wrist & \{right,left\}\_wrist \\
    \{r,l\}hip & \{right,left\}\_hip \\    
    \{r,l\}knee & \{right,left\}\_knee \\    
    \{r,l\}ankle & \{right,left\}\_ankle \\            
    pelvis & (left\_hip + right\_hip) / 2 \\
    clavicle & (left\_shoulder + right\_shoulder) / 2 \\    
    waist & (pelvis + clavicle) / 2 \\
    neck & (mouth\_left + clavicle) / 2\\
    thorax   & (pelvis + clavicle) / 2 \\
    \bottomrule
  \end{tabular}
  \caption{Approximate mapping of BlazePose landmark points to Vicon joints.}
  \label{tab:Blaze2Vicon}
\end{table}

\end{document}